\documentclass{ieeetj}
\usepackage{amsmath,amsfonts,amssymb}
\usepackage{mathrsfs}
\usepackage{mathtools}
\usepackage{dsfont}
\usepackage{bm}
\usepackage[long]{optidef}

\usepackage{ifthen}
\newboolean{mathfamilysansserif}
\setboolean{mathfamilysansserif}{false} 
\ifthenelse{\boolean{mathfamilysansserif}}{%
	\newcommand*{\upm}[1]{\mathsf{#1}}%
}{
	\newcommand*{\upm}[1]{\mathrm{#1}}%
}

\DeclarePairedDelimiter{\pdel}{(}{)} %

\DeclarePairedDelimiter{\abs}{\lvert}{\rvert}

\DeclarePairedDelimiter{\bmat}{[}{]} %
\DeclarePairedDelimiter{\uset}{\{}{\}} %
\DeclarePairedDelimiter{\oset}{(}{)} %
\DeclareMathOperator*{\argmin}{\arg\min}

\newcommand{\eul}{{\upm{e}}}
\newcommand*{\onevec}{\bm{1}}
\newcommand*{\nullvec}{\bm{0}}

\newcommand*{\idmat}{\bm{I}}

\newcommand*\Nset{\mathbb{N}}

\newcommand*\Rset{\mathbb{R}}

\DeclarePairedDelimiter{\norm}{\lVert}{\rVert}

\newcommand*{\T}{\upm{T}}
\newcommand*{\frobnorm}[1]{\norm*{#1}_\upm{F}}

\newcommand*{\trace}{\operatorname{Tr}\pdel*}

\newcommand*{\Diag}{\operatorname{Diag}\pdel*}

\NewDocumentCommand{\expect}{o}{
	\IfNoValueTF{#1}{\mathbb{E}\pdel*}
	{\operatorname{E}_{#1}\pdel*}
}
\NewDocumentCommand{\emexpect}{o}{
	\IfNoValueTF{#1}{\widehat{\mathbb{E}}\pdel*}
	{\widehat{\mathbb{E}}_{#1}\pdel*}
}
\NewDocumentCommand{\var}{o}{
	\IfNoValueTF{#1}{\operatorname{Var}\pdel*}
	{\operatorname{E}_{#1}\pdel*}
}
\NewDocumentCommand{\emvar}{o}{
	\IfNoValueTF{#1}{\widehat{\operatorname{Var}}\pdel*}
	{\widehat{\operatorname{Var}}_{#1}\pdel*}
}

\NewDocumentCommand{\jcby}{o}{\IfNoValueTF{#1}{\operatorname{D}\!}{\operatorname{D}_{#1}\!}} %

\newcommand{\cmdstkout}[1]{\ifmmode\upm{\sout{\ensuremath{#1}}}\else\sout{#1}\fi}%

\usepackage{listings}
\lstdefinelanguage{NTSpython}[]{Python}{
	deletekeywords=[2]{max, range, sum, abs}, %
	morekeywords={as}
}
\usepackage{xspace}
\usepackage{glossaries}
\usepackage{siunitx}
\usepackage{hyperref}

\usepackage{tikz}
\usepackage{tikzscale}
\usepackage{pgfplots}
\usepgfplotslibrary{fillbetween}
\usepackage{tabularx}
\usepackage{multirow}
\usepackage{booktabs}
\usepackage{algorithm}
\usepackage{algpseudocode}
\usepackage{amsthm}
\usepackage{colortbl}
\usepackage{diagbox}
\usepackage{makecell}

\usetikzlibrary{external}
\usetikzlibrary{pgfplots.groupplots} %
\usetikzlibrary{positioning,intersections,fit,shapes,arrows.meta}
\usetikzlibrary{backgrounds}
\usetikzlibrary{calc}
\usetikzlibrary{spy}
\usetikzlibrary{plotmarks}

\definecolor{red}{HTML}{cc2900}
\definecolor{blue}{HTML}{0E84B2}
\definecolor{green}{HTML}{3cb44b}
\definecolor{orange}{HTML}{f58231}
\definecolor{purple}{HTML}{984ea3}
\definecolor{cyan}{HTML}{00FFFF}

\definecolor{thisPurple}{rgb}{0.502,0.501,0.969}
\definecolor{veryDarkGray}{rgb}{0.2,0.2,0.2}
\definecolor{darkGray}{rgb}{0.563,0.563,0.563}
\definecolor{midGray}{rgb}{0.783,0.783,0.783}
\definecolor{lightGray}{rgb}{0.902,0.902,0.902}
\definecolor{lightBlue}{rgb}{0.75,0.85,1.0}
\definecolor{darkBlue}{rgb}{0.45,0.55,1.0}
\definecolor{darkRed}{rgb}{0.961,0.426,0.411}
\definecolor{lightRed}{rgb}{0.99,0.626,0.611}
\definecolor{darkGreen}{rgb}{0.426,0.761,0.411}

\definecolor{green1}{HTML}{bae4b3}
\definecolor{green2}{HTML}{74c476}
\definecolor{green3}{HTML}{31a354}
\definecolor{green4}{HTML}{006d2c}
\definecolor{blue1}{HTML}{accbff}
\definecolor{blue2}{HTML}{92bbff}
\definecolor{blue3}{HTML}{78aaff}
\definecolor{blue4}{HTML}{649eff}
\definecolor{blue5}{HTML}{4188ff}
\definecolor{red1}{HTML}{ffbaba}
\definecolor{red2}{HTML}{ff7b7b}
\definecolor{red3}{HTML}{ff5252}
\definecolor{red4}{HTML}{ff0000}
\definecolor{red5}{HTML}{a70000}

\theoremstyle{plain}%
\newtheorem{thm}{Theorem}[section]

\newtheorem{prop}[thm]{Proposition}

\theoremstyle{remark}
\newtheorem*{rem}{Remark}

\pgfplotscreateplotcyclelist{markcycle}{mark = o, mark = triangle, mark = diamond, mark = square, mark = pentagon, mark = star, mark = oplus, mark = otimes}

\pgfplotsset{
	minor grid style={dotted},
	major grid style={solid},
	grid = major, 
	cycle list name = markcycle,
	every axis plot/.append style={thick, mark options=solid,},
	legend style={at={(1,1.03)}, draw=none, fill=none, font=\small, anchor=south east, /tikz/every even column/.append style={column sep=0.2cm}},
}

\newcommand\offs{0.1} %
\tikzset{
	nc/.style={circle, draw=veryDarkGray, fill=lightGray, minimum width=7mm, inner sep=0.00cm, align=center, font=\footnotesize\sffamily},
	nst/.style={draw=none, align=center, font=\footnotesize},
	>={Latex[scale=0.6]},
	nr/.style={draw, rectangle, fill=lightGray, minimum height=6mm, minimum width=6mm, inner sep=1mm, rounded corners=0mm, align=center},
	nl/.style={draw=none, rectangle, fill=lightBlue, opacity=0.5, inner sep=3mm, rounded corners=2mm, align=center},
	bdot/.style={draw, circle, fill=veryDarkGray, minimum width=1mm, inner sep=0mm, align=center},
	circgray/.style={draw, circle, fill=lightGray, minimum width=7mm, inner sep=0mm, align=center},
	circred/.style={draw, circle, fill=lightRed, minimum width=7mm, inner sep=0mm, align=center, font=\footnotesize},
	txt/.style={inner sep=0.5mm, align=center, font=\footnotesize},
	lsty/.style={line width=0.4mm, veryDarkGray},
	Mb/.style={nr, minimum width=7mm, minimum height=7mm},
	Pb/.style={nr},
	Qb/.style={nr},
	Ab/.style={nr},
}

\makeatletter
\DeclareRobustCommand\onedot{\futurelet\@let@token\@onedot}
\def\@onedot{\ifx\@let@token.\else.\null\fi\xspace}

\def\eg{e.g\onedot} 
\def\ie{i.e\onedot}

\def\wrt{w.r.t\onedot} 

\makeatother

\newcommand*{\itidx}[1]{^{(#1)}}

\newcommand*{\vct}[1]{\bm{#1}}
\newcommand*{\mtr}[1]{\bm{#1}} 
\newcommand*{\ts}[1]{\bm{\mathcal{#1}}}  %
\newcommand*{\unf}[1]{_{(#1)}}  %
\newcommand*{\idx}[1]{^{(#1)}} 
\newcommand*{\matel}[2]{\bmat*{#1}_{#2}}
\newcommand*{\cpd}[1]{\left[\!\left[{#1}\right]\!\right]}

\newcommand*{\hprod}{\odot}
\newcommand*{\hdiv}{\oslash}
\newcommand*{\krprod}{\circledast}
\newcommand*{\toprod}{\circ}
\newcommand*{\tmprod}[1]{{\times\mkern-1.5mu}_{#1}}

\newcommand*{\sfttresh}[2]{S_{#1}\left(#2\right)}

\newcommand*{\vh}{\vct{h}}
\newcommand*{\vp}{\vct{p}}
\newcommand*{\bs}{\vct{s}}

\newcommand*{\vx}{\vct{x}}

\newcommand*{\bA}{\mtr{A}}
\newcommand*{\bB}{\mtr{B}}
\newcommand*{\bC}{\mtr{C}}

\newcommand*{\bN}{\mtr{N}}
\newcommand*{\bO}{\mtr{O}}
\newcommand*{\bP}{\mtr{P}}
\newcommand*{\bQ}{\mtr{Q}}
\newcommand*{\bR}{\mtr{R}}

\newcommand*{\bX}{\mtr{X}}
\newcommand*{\bY}{\mtr{Y}}
\newcommand*{\bZ}{\mtr{Z}}

\newcommand*{\tA}{\ts{A}}

\newcommand*{\tE}{\ts{E}}

\newcommand*{\tH}{\ts{H}}
\newcommand*{\tM}{\ts{M}}
\newcommand*{\tN}{\ts{N}}
\newcommand*{\tO}{\ts{O}}

\newcommand*{\tR}{\ts{R}}
\newcommand*{\tS}{\ts{S}}

\newcommand*{\tW}{\ts{W}}
\newcommand*{\tX}{\ts{X}}
\newcommand*{\tY}{\ts{Y}}
\newcommand*{\tZ}{\ts{Z}}

\newcommand*{\hbA}{\widehat{\bA}}
\newcommand*{\hbX}{\widehat{\bX}}
\newcommand*{\ttX}{\widetilde{\tX}}
\newcommand*{\ttO}{\widetilde{\tO}}

\newcommand*{\btheta}{\vct{\theta}}

\newcommand*{\onetens}{\bm{1}}

\newcommand*{\ocplx}{\mathcal{O}\pdel}

\newcommand*{\auc}{\operatorname{AUC}}

\newcommand*{\sauc}{\operatorname{sAUC}}

\newcommand*{\mr}[1]{\mathrm{#1}}

\newcommand*{\anoamp}{A_{\mr{ano}}}
\newcommand*{\anoprob}{p_{\mr{ano}}}
\newcommand*{\noisestd}{\sigma_{\mr{noise}}}

\newcommand*{\obsprob}{p_{\mr{obs}}}

\newcommand*{\cprank}{R_{\mr{cpd}}}

\newcommand*{\model}{\mathcal{F}}
\newcommand*{\modelly}{\breve{\mathcal{F}}}

\newcommand*{\ssample}{\mathcal{S}}
\newcommand*{\dataset}{\mathcal{D}}

\newcommand*{\obsfun}{\omega}
\newcommand*{\featvec}{\vh} %
\newcommand*{\embedfun}{f_\mr{ft}} %
\newcommand*{\embedfuni}[1]{f_{\mr{ft,#1}}} %
\newcommand*{\parfun}{f_{\mr{par}}} %
\newcommand*{\parfuni}[1]{f_{\mr{par,#1}}} %
\newcommand*{\stepfun}{u}
\newcommand*{\logistic}{\operatorname{S}}
\newcommand*{\Rgt}{R_\mr{gt}}
\newcommand*{\setAgt}{\mathcal{A}^\mr{gt}}
\newcommand*{\setestAgt}{\mathcal{A}^\mr{la}}
\newcommand*{\setAul}{\mathcal{A}^\mr{err}}

\newcommand*{\bsca}{{BSCA-AD}}
\newcommand*{\aubsca}{{AU-BSCA}}
\newcommand*{\bbcd}{{BBCD}}
\newcommand*{\tbsca}{{tBSCA-AD}}
\newcommand*{\tbscarlx}{{tBSCA-AD-AUG}}
\newcommand*{\utbsca}{U-tBSCA}
\newcommand*{\umbscarlx}{U-mBSCA-AUG}
\newcommand*{\utbscarlx}{U-tBSCA-AUG}
\newcommand*{\aumbscarlx}{AU-mBSCA-AUG}
\newcommand*{\autbsca}{AU-tBSCA-AUG}
\newcommand*{\kasai}{{HTO-AD}}

\newacronym{ad}{AD}{anomaly detection}
\newacronym{roc}{ROC}{receiver operating characteristic}
\newacronym{auc}{AUC}{area under the curve}

\newacronym{cpd}{CPD}{canonical polyadic decomposition}
\newacronym{svd}{SVD}{singular value decomposition}
\newacronym{pca}{PCA}{principal component analysis}
\newacronym{rpca}{RPCA}{robust principal component analysis}
\newacronym{bcd}{BCD}{block coordinate descent}
\newacronym{bsca}{BSCA}{block successive convex approximation}

\newacronym{dn}{DN}{deep network}
\newacronym{nn}{NN}{neural network}
\newacronym{lista}{LISTA}{robust principal component analysis}
\newacronym{mlp}{MLP}{multilayer perceptron}
\newacronym{sgd}{SGD}{stochastic gradient descent}

\def\BibTeX{{\rm B\kern-.05em{\sc i\kern-.025em b}\kern-.08em
		T\kern-.1667em\lower.7ex\hbox{E}\kern-.125emX}}
\AtBeginDocument{\definecolor{tmlcncolor}{cmyk}{0.93,0.59,0.15,0.02}\definecolor{NavyBlue}{RGB}{0,86,125}}

\def\OJlogo{}
\def\seclogo{}
\def\authorrefmark#1{\ensuremath{^{\textbf{#1}}}}

\begin{document}
	\receiveddate{19 September, 2024}
	\reviseddate{18 February, 2025}
	\accepteddate{27 February, 2025}
	\publisheddate{10 March, 2025}
	\jvol{6}
	\pubyear{2025}
	
	\markboth{Adaptive Anomaly Detection in Network Flows with Low-Rank Tensor Decompositions and Deep Unrolling}{Schynol and Pesavento}
	
	\title{Adaptive Anomaly Detection in Network Flows with Low-Rank Tensor Decompositions and Deep Unrolling}
	
	\author{~Lukas Schynol \authorrefmark{1} (Graduate Student Member, IEEE) and Marius Pesavento\authorrefmark{1} (Senior Member, IEEE)}
	\affil{Technische Universit\"at Darmstadt, Darmstadt, 64283 Germany}
	\corresp{Corresponding author: Lukas Schynol (email: lschynol@nt.tu-darmstadt.de).}
	\authornote{This work was financially supported by the Federal Ministry of Education and Research of Germany in the project "Open6GHub" (grant no. 16KISK014). Parts of this work have been presented at the 9th International Workshop on Computational Advances in Multi-Sensor Adaptive Processing, CAMSAP 2023.}

	\begin{abstract}
		\Gls{ad} is increasingly recognized as a key component for ensuring the resilience of future communication systems.
		While deep learning has shown state-of-the-art \gls{ad} performance, its application in critical systems is hindered by concerns regarding training data efficiency, domain adaptation and interpretability.
		This work considers \gls{ad} in network flows using incomplete measurements, leveraging a robust tensor decomposition approach and deep unrolling techniques to address these challenges.
		We first propose a novel block-successive convex approximation algorithm based on a regularized model-fitting objective where the normal flows are modeled as low-rank tensors and anomalies as sparse.
		An augmentation of the objective is introduced to decrease the computational cost.
		We apply deep unrolling to derive a novel deep network architecture based on our proposed algorithm, treating the regularization parameters as learnable weights.
		Inspired by Bayesian approaches, we extend the model architecture to perform online adaptation to per-flow and per-time-step statistics, improving \gls{ad} performance while maintaining a low parameter count and preserving the problem's permutation equivariances.
		To optimize the deep network weights for detection performance, we employ a homotopy optimization approach based on an efficient approximation of the area under the receiver operating characteristic curve.
		Extensive experiments on synthetic and real-world data demonstrate that our proposed deep network architecture exhibits a high training data efficiency, outperforms reference methods, and adapts seamlessly to varying network topologies.
	\end{abstract}
	\glsresetall
	
	\maketitle
	\section{Introduction}
	\Gls{ad} plays a significant role in the active resilience of communication systems \cite{sterbenzResilienceSurvivabilityCommunication2010} by enabling targeted remediation and recovery with the detection of undesired system states.
	For instance, traffic anomalies in communication networks indicate attacks, intrusions, routing errors and equipment failures \cite{bhuyanNetworkAnomalyDetection2014, fernandesComprehensiveSurveyNetwork2019}, whereas anomalies in the transient power spectral density of radio bands of wireless networks may indicate unauthorized usage, adversarial jamming or outages \cite{liuALDOAnomalyDetection2009, zhangByzantineAttackDefense2015}.
	Due to the increasing reliance on wireless systems and their permeation into safety-critical applications, \gls{ad} will not only play the role of an ad-hoc feature in future wireless communication standards, but become a central design component \cite{shafinArtificialIntelligenceEnabledCellular2020}.
	As such, \gls{ad} is an active field of research \cite{chandolaAnomalyDetectionSurvey2009,fernandesComprehensiveSurveyNetwork2019}.

	\par
	A particular optimization-guided and versatile approach for \gls{ad} is \gls{rpca}, first analyzed in \cite{candesRobustPrincipalComponent2011}.
	Given an incomplete measurement matrix $\bY$, \gls{rpca} recovers a low-rank matrix $\bX$ and a sparse error matrix $\bA$ from $\bY$.
	A variation on \gls{rpca} is proposed in \cite{mardaniDynamicAnomalographyTracking2013} to perform traffic recovery and traffic \gls{ad} in networks using only cumulative traffic measurements acquired at the network links.
	Thereby, normal traffic flows are modeled as a low-rank matrix component, while anomalies are represented as sparse perturbations of the traffic observed at the links.
	The \gls{rpca}-based \gls{ad} approach is remarkably robust and flexible due to its capability to reconstruct traffic while relying only on incomplete measurements \cite{kasaiNetworkVolumeAnomaly2016}.
	Similar \gls{rpca}-based methods are applied to \gls{ad} in monitoring of urban vehicular traffic \cite{sofuogluGLOSSTensorBasedAnomaly2022}, wireless spectra \cite{dingRobustOnlineSpectrum2017, liRobustOnlinePrediction2022} and power systems \cite{gaoIdentificationSuccessiveUnobservable2016}.
	The notion of low-rank matrices can be extended to low-rank tensors to better model additional structure such as periodicity of the normal data \cite{tanTrafficVolumeData2013, bazerqueRankRegularizationBayesian2013}.
	\par
	Deep learning has gained significant attention in signal processing due to its state-of-the-art performance in fields such as computer vision, natural language processing and \gls{ad} \cite{chalapathyDeepLearningAnomaly2021}.
	However, generic \gls{dn} architectures such as fully connected \glspl{nn}, convolutional \glspl{nn} or long-short-term memory require a substantial amount of training data to generalize.
	This is a problem for \gls{ad} in communication systems, where real-world data is a scarce resource \cite{shafinArtificialIntelligenceEnabledCellular2020}.
	Moreover, generic \glspl{nn} are considered as black boxes, potentially excluding them from an application in trustworthy critical systems \cite{arrietaExplainableArtificialIntelligence2020}.
	The concept of deep unrolling and model-aided deep learning tackles these issues by bridging the gap between generic \gls{nn} architectures and traditional signal processing algorithms using domain-knowledge-based statistical or physical models \cite{mongaAlgorithmUnrollingInterpretable2021, shlezingerModelBasedDeepLearning2022a, gregorLearningFastApproximations2010}.
	Deep unrolling builds on classical optimization and algorithm design:
	Algorithm iterations are modified into \gls{dn} layers with learnable weights, which are then learned through empirical risk minimization.
	This approach decouples the algorithm objective and the task objective, allowing for an algorithm objective which is feasible to handle, e.g., a likelihood, while the potentially challenging task objective is used to tune the learnable weights.
	Compared to \gls{nn} architectures with little integrated domain knowledge, the resulting model architectures usually exhibit significantly fewer learnable weights and, therefore, achieve remarkable training data efficiency, generalization capability and improved explainability, while retaining the advantages in performance and computational cost \cite{mongaAlgorithmUnrollingInterpretable2021, shlezingerModelBasedDeepLearning2022a, schynolCoordinatedSumRateMaximization2023}.
	\par
	Deep unrolling has previously been considered in the context of \gls{rpca} and low-rank tensor decomposition for diverse image processing tasks such as foreground-background-separation, image denoising and artifact removal \cite{solomonDeepUnfoldedRobust2020, cohenDeepConvolutionalRobust2019, vanluongDeepUnfoldedReferenceBasedRPCA2021, maiDeepUnrolledLowRank2022, joukovskyInterpretableNeuralNetworks2024, dongDeepUnfoldedTensor2023, caiLearnedRobustPCA2021, miriyathanthrigeDeepUnfoldingIteratively2022}.
	These works primarily focus on the recovery of the normal or anomalous data, which is much different from the anomaly \textit{detection} task.
	\par
	In this work, we address the issue of \gls{ad} in network flows from link loads, which poses a particular challenge to common \gls{dn} approaches.
	First, since link loads comprise of linear combinations of network flows, the solution space is generally a higher-dimensional space than the input data space, preventing the straightforward application of convolutional \glspl{nn} or graph \glspl{nn}.
	Secondly, only an incomplete observation of the link loads may be available. Thirdly, compounding on the general scarcity of training data in \gls{ad} tasks, networks are constantly evolving due to nodes entering or leaving and changes in the routing of network flows.
	\par
	We are thus motivated to follow a model-aided approach and propose a novel adaptive \gls{dn} architecture based on deep unrolling.
	To summarize, our contributions are:
	\begin{itemize}
 		\item We propose a novel algorithm for network flow recovery and \gls{ad} from incomplete link load measurements, which models normal flows as a low-rank \gls{cpd} and anomalies as sparse.
		\item An augmentation of the optimization objective is introduced to increase the degrees of freedom and substantially reduce the computational cost.
		\item The algorithms are unfolded into \gls{dn} architectures, enabling the capability to process incomplete measurements while seamlessly adapting to changing network topologies.
		In addition, we propose a novel architecture which adapts to the statistical properties of individual network flows and time steps by utilizing the tensor structure, while retaining the invariance properties of the original problem. 
		\item Instead of risk minimization based on the $\ell_1$-distance or $\ell_2$-distance between estimated and ground-truth signals as in \cite{solomonDeepUnfoldedRobust2020, cohenDeepConvolutionalRobust2019, vanluongDeepUnfoldedReferenceBasedRPCA2021, maiDeepUnrolledLowRank2022, joukovskyInterpretableNeuralNetworks2024, dongDeepUnfoldedTensor2023, caiLearnedRobustPCA2021, miriyathanthrigeDeepUnfoldingIteratively2022}, we propose a homotopy optimization scheme to tune model weights based on the \gls{auc} of the \gls{roc}, striking a balance between maximizing the probability of detection and minimizing the probability of false alarms.
		\item We approximate the corresponding loss function to reduce the computational cost during training.
		\item Extensive simulations on synthetic and real-world data characterize the proposed algorithms and architectures, validate their data efficiency and adaptivity, and compare their performance them against reference methods.
	\end{itemize}
	This works extends our preliminary work in \cite{schynolDeepUnrollingAnomaly2023} in several aspects, in particular, by considering a tensor-based data model over a matrix factorization model, by proposing objective augmentation to improve computational efficiency, by proposing a parameter adaptation scheme with increased degrees of freedom, by proposing a computationally more efficient loss function and by significantly expanding the empirical verification.
	\subsection{Paper Outline}
	The remainder of the paper is organized as follows: In Section \ref{sec:relwork}, \gls{ad} based on network flows with particular focus on \gls{rpca}-based and low-rank tensor methods is reviewed. 
	Sec.~\ref{sec:problemdef} introduces the system model. %
	The proposed tensor-based recovery algorithm and its augmentation are presented in Sec.~\ref{sec:bsca_algs}. 
	In Sec.~\ref{sec:unroll_tbsca}, we introduce a novel \gls{dn} architecture inspired by the recovery algorithms and deep unrolling techniques, then extend it to adapt dynamically to network flow statistics.
	The proposed methods are evaluated via extensive simulations in Sec.~\ref{sec:simulations}.
	We summarize and provide an outlook in Sec.~\ref{sec:conclusion}.
	\subsection{Notation} \label{sec:notation}
	\begin{table}
		\caption{Notation conventions.\label{tab:notation}}
		\begin{tabularx}{\linewidth}{cl}
			\toprule
			$x$, $\vx$, $\bX$, $\tX$, $\mathcal{X}$ & scalar, vector, matrix, tensor, set\\
			$\left[\vp_i| i=1,\dots, I\right]$ & concatenation of $(\vp_i)_{i=1}^I$ into a matrix\\
			$\matel{\tX}{i,j,k}$ & $(i, j, k)$th element of $\tX$\\
			$\matel{\tX}{i,:,:}$ / $\matel{\tX}{:,j,:}$ / $\matel{\tX}{:,:,k}$ & tensor slices along $1$st / $2$nd / $3$rd mode\\
			$\matel{\tX}{\Sigma}$ & sum of elements $\sum_{i, j, k} \matel{\tX}{i, j, k}$\\
			$\Diag{\vx}$ & diagonal matrix with the entries of $\vx$\\
			$\tX\unf{n}$ & mode-$n$ unfolding\\
			$\hprod$ / $\hdiv$ & Hadamard multiplication / division\\
			$\tX^2$ & $\tX \hprod \tX$\\
			$\krprod$ & Khatri-Rao product\\
			$\toprod$ & outer product / function composition\\
			$\tmprod{n}$  & $n$-mode tensor-matrix product\\
			$\sfttresh{\mu}{\cdot}$ & soft-thresholding with threshold $\mu$\\
			\bottomrule
		\end{tabularx}
	\end{table}
	Important notational conventions are summarized in Tab.~\ref{tab:notation}.
	Indexing and slicing conventions translate to vectors and matrices, e.g., $\matel{\bX}{i, :}$ is the vector representing the $i$th row of $\bX$.
	For matrices $\bA\in\Rset^{I \times R}$, $\bB\in\Rset^{J \times R}$ and $\bC\in\Rset^{I \times R}$:
	\begin{align}
		\cpd{\bA,  \bB, \bC} = \sum_{r=1}^{R} \matel{\bA}{:, r} \circ \matel{\bB}{:, r} \circ \matel{\bC}{:, r} \in \Rset^{I \times J \times K} \nonumber
	\end{align}
	is a rank-$R$ \gls{cpd} tensor model.
	Furthermore, we define the entry-wise $\ell_1$-norm $\norm{\bX}_1 = \sum_{i, j} \abs{\matel{\bX}{i,j}}$ and the Frobenius norm $\frobnorm{\bX} = (\sum_{i, j} \abs{\matel{\bX}{i,j}}^2)^{1/2}$ of $\bX$, which are similarly extended to tensors. 
	Sample mean and variances based on the elements of a tensor $\tX$ are denoted as $\emexpect{\tX}$ and $\emvar{\tX}$.
	Lastly, given tensors $\tO \in \uset{0, 1}^{I\times J \times K}$, $\emexpect[\tO]{\tX}$ and $\emvar[\tO]{\tX}$ are the masked sample mean and variance
	\begin{align*}
		\emexpect[\tO]{\tX} &=\matel{\tO \hprod \tX}{\Sigma} \;\big/ \matel{\tO}{\Sigma},\\
		\emvar[\tO]{\tX} &= \sum_{i, j, k} \left(\matel{\tO \hprod \tX}{i, j, k} - \emexpect[\tO]{\tX}  \right)^2 \big/  \big(\matel{\tO}{\Sigma} - 1\big),
	\end{align*}
	where $[\cdot]_\Sigma$ is defined as in Tab.~\ref{tab:notation}.
	The reader is referred to \cite{koldaTensorDecompositionsApplications2009} for an introduction to tensor processing.
	\section{Related Work}\label{sec:relwork}
	\gls{ad} in general and in particular \gls{ad} in network traffic has been widely investigated \cite{chandolaAnomalyDetectionSurvey2009, fernandesComprehensiveSurveyNetwork2019}.
	Considering classical model-driven approaches, early work performing \gls{ad} in network traffic from link load measurements uses \acrlong{pca} \cite{lakhinaDiagnosingNetworkwideTraffic2004}.
	\Gls{rpca}-type \gls{ad} algorithms can be broadly categorized into batch methods and online methods, while being either matrix-based or tensor-based.
	The authors of \cite{mardaniDynamicAnomalographyTracking2013} were first to propose \gls{rpca}-based \cite{candesRobustPrincipalComponent2011} batch and online algorithms for \gls{ad} in network traffic from incomplete link measurements.
	The considered system model and recovery problem is adapted in our work.
	An accelerated and parallelizable algorithm based on successive convex approximations for complete measurements is developed in \cite{yangInexactBlockCoordinate2020}.
	The work in \cite{mardaniEstimatingTrafficAnomaly2016} incorporates additional direct observations of point-to-point flows into the data model of \cite{mardaniDynamicAnomalographyTracking2013}, and utilizes additional assumptions on the correlation of network flows over time.
	The work in \cite{zhangStructureRegularizedTraffic2016} builds on the online algorithm in \cite{mardaniDynamicAnomalographyTracking2013}, but replaces the low-rank assumption of the normal traffic with structure regularization while using complete link observations.
	Ye et al. \cite{yeAnomalyTolerantTrafficMatrix2017} extend the approach in \cite{mardaniEstimatingTrafficAnomaly2016} with similar structure regularizations.
	\par
	Several works extended \gls{rpca}-based \gls{ad} from matrix to tensor methods, which can improve the recovery or \gls{ad} performance at the cost of computational complexity.
	In \cite{kasaiNetworkVolumeAnomaly2016}, an online \gls{ad} algorithm is developed by integrating additional structural information about network traffic compared to \cite{mardaniDynamicAnomalographyTracking2013} with the extension to 3D-tensor \gls{rpca}.
	The authors of \cite{xieFastTensorFactorization2017} and \cite{wangHankelstructuredTensorRobust2021} propose a batch \gls{ad} algorithm by following a similar tensor-based approach as \cite{kasaiNetworkVolumeAnomaly2016}, however, fully observed uncompressed point-to-point flows are considered as input data.
	In \cite{wangAnomalyAwareNetworkTraffic2020}, this is extended to incomplete measurements.
	A tensor-based batch algorithm that accommodates either compressed or incomplete flows is proposed in \cite{yuSLRTASparseLowrank2021}.

	Deep unrolling has first been explicitly applied in solving the LASSO problem \cite{gregorLearningFastApproximations2010}.
	It has since been considered for \gls{rpca}-type algorithms, primarily focusing on image processing applications \cite{cohenDeepConvolutionalRobust2019, solomonDeepUnfoldedRobust2020, vanluongDeepUnfoldedReferenceBasedRPCA2021, maiDeepUnrolledLowRank2022, joukovskyInterpretableNeuralNetworks2024, dongDeepUnfoldedTensor2023}.
	The authors of \cite{cohenDeepConvolutionalRobust2019} propose a scalable \gls{rpca}-based algorithm by unrolling the scaled gradient descend procedure for the recovery of low-rank images.
	In \cite{solomonDeepUnfoldedRobust2020}, the authors unroll an algorithm for clutter suppression in images derived from a fixed-point analysis.
	The resulting method requires a costly \gls{svd} in each iteration, and requires a complete observation of the data.
	The work in \cite{vanluongDeepUnfoldedReferenceBasedRPCA2021} extends \cite{solomonDeepUnfoldedRobust2020} by incorporating knowledge about the correlation between the elements of the sparse matrix, improving reconstructive performance.
	The authors of \cite{joukovskyInterpretableNeuralNetworks2024} perform foreground-background separation by unrolling a masked \gls{rpca} algorithm, which directly seeks a mask to separate outliers from the low-rank image data.
	Interestingly, their proposed loss function partially consists of a cross-entropy term of the mask, which is enabled by the assumption that the mask only takes binary values.
	In \cite{maiDeepUnrolledLowRank2022}, an extension of \gls{rpca} to 3D-tensors is used to perform robust image completion.
	Recently in \cite{dongDeepUnfoldedTensor2023}, tensor-\gls{rpca} with scaled gradient descent is unrolled to recover the low-rank components of images in a semi-supervised manner.
	Applications of \gls{rpca}-based deep unrolling to artifact removal in radar applications similar to \cite{solomonDeepUnfoldedRobust2020} are considered in \cite{miriyathanthrigeDeepUnfoldingIteratively2022}.
	Overall, all these works primarily optimize a reconstruction objective, but do not optimize the task of \gls{ad} performance specifically.
	Regarding works based on low-rank tensor decompositions, we note that only the work in \cite{kasaiNetworkVolumeAnomaly2016} combines incomplete observations, anomalies and flow compression.
	\section{System Model}\label{sec:problemdef}\label{sec:sysmodel}
	\begin{figure}
		\centering
\tikzsetnextfilename{main_network_model}
\begin{tikzpicture}[
	]
	
	\node[circgray] (n1) at (0, 0) {1};
	\node[circgray] (n2) at (0, 2.5) {2};
	\node[circgray] (n3) at (2.5, 0) {3};
	\node[circgray] (n4) at (2.5, 2.5) {4};
	\draw[lsty] (n1) to (n2);
	\draw[lsty] (n1) to (n3);
	\draw[lsty] (n2) to (n3);
	\draw[lsty] (n2) to (n4);
	\node[nst] at (1.45, 3.05) {$y_{2\to 4} = \textcolor{darkBlue}{z_1 + a_1} +  \textcolor{darkRed}{z_3 + a_3}$};
	\node[nst] at (-0.8, 1.25) {$y_{1\to 2} =$\\$ \textcolor{darkBlue}{z_1 + a_1}$};
	\node[nst, rotate=-45] at (1.45, 1.45) {$y_{2\to 3} = \textcolor{darkGreen}{z_2 + a_2}$};
	\node[nst, rotate=-45] at (1.05, 1.05) {$y_{2\gets 3} = \textcolor{darkRed}{z_3 + a_3}$};
	
	\begin{scope}[on background layer]
		\draw[lsty, darkBlue, ->] ([shift={(-0.10, -0.05)}]n1.north) to ([shift={(-0.10, 0.0)}]n2.mid); %
		\draw[lsty, darkBlue, ->] ([shift={(-0.05, 0.10)}]n2.east) to ([shift={(0.05, 0.10)}]n4.west); %
		\draw[lsty, darkRed, -] ([shift={(-0.08, -0.06)}]n2.{south east}) to ([shift={(-0.06, -0.08)}]n3.{north west}); %

		\draw[lsty, darkRed, ->] ([shift={(-0.05, -0.10)}]n2.east) to ([shift={(0.05, -0.10)}]n4.west); %
		\draw[lsty, darkGreen, <-] ([shift={(0.06, 0.08)}]n3.{north west}) to ([shift={(0.07, 0.07)}]n2.{south east}); %

	\end{scope}
	\draw[lsty, darkBlue, opacity=0.5] ([shift={(-0.10, 0.0)}]n2.south) to[out=90, in=180] ([shift={(-0.01, 0.10)}]n2.east); %
	\draw[lsty, darkRed, opacity=0.5] ([shift={(-0.08, -0.06)}]n2.{south east}) to[out=135, in=180] ([shift={(-0.02, -0.10)}]n2.east); %

\end{tikzpicture}
\vspace{-1mm}
		\caption{Example network with $N=4$ nodes and $F=3$ directed flows routed across the edges. The directed edge traffic $y_{2\to 4}$ from node $2$ to $4$ is a superposition of the directed flow from node $1$ to $4$ (blue) and from node $3$ to $4$ (red).}
		\label{fig:routing}
	\end{figure}
	Communication networks are commonly modeled as graphs, where each traffic flow from one node to the other passes over certain edges.
	To be specific, consider a directed graph $\mathcal{G}=\oset{\mathcal{V}, \mathcal{E}}$ with a set of $N$ nodes $\mathcal{V}$ and a set of $E$ directed edges or links $\mathcal{E}$, which are indexed by $j=1, \dots, E$.
	At most $F=N(N-1)$ flows in total are routed between the nodes along the directed edges.
	For each time step $t$ for $t=1,\dots,T$, the $i$th flow for $i=1,\dots, F$ is a superposition of a normal flow $\matel{\bZ}{i,t}$ and an anomalous flow $\matel{\bA}{i,t}$.
	The matrices $\bZ \in \Rset^{F\times T}$ and $\bA \in \Rset^{F\times T}$ summarize all normal and anomalous flows across all time steps, respectively.
	Assuming a time-invariant routing matrix $\bR \in \uset{0, 1}^{E\times F}$, its elements $\matel{\bR}{j, i}$ indicate that the $i$th flow is routed along the $j$th link.
	The $j$th link load at time $t$ is consequently $\sum_{i=1}^{F}\matel{\bR}{j, i} (\matel{\bZ}{i,t} + \matel{\bA}{i,t})$.
	The routing of flows in a simple network example is illustrated in Fig.~\ref{fig:routing}.
	Note that although for simplicity the routing matrix elements are limited to $0$ or $1$ in this work, which essentially implies single-path routing, all of the algorithms considered in this work are fundamentally applicable for arbitrary routing matrices $\bR\in\Rset^{E\times F}$.
	The measurement matrix $\bY$, where $\matel{\bY}{j,t}$ is the $j$th link traffic at time $t$, is defined by the model
	\vspace{-1mm}
	\begin{equation}
		\bY = \bO \odot \left( \bR \left( \bZ + \bA \right) + \bN \right). \label{eq:obs_model}
	\end{equation}
	\vspace{-1mm}
	Here, the observation matrix $\bO \in \uset{0,1}^{E\times T}$ indicates whether a measurement is observed ($\matel{\bO}{j,t} = 1$) or missing ($\matel{\bO}{j,t} = 0$) at a particular link $j$ and time instance $t$, and the noise matrix $\bN\in\Rset^{E\times T}$ models additional measurement noise or modeling errors.
	Given \eqref{eq:obs_model}, a scenario realization is fully defined by the tuple $\ssample = \oset{\bO, \bR, \bZ, \bA, \bN}$.
	\par 
	The model in \eqref{eq:obs_model} has broad applicability.
	For instance, the nodes $\mathcal{V}$ could describe routers in an internet backbone network \cite{mardaniDynamicAnomalographyTracking2013}, where anomalous flows can indicate faults and adversarial attacks.
	In this case, the model is particularly attractive since internet traffic is easier to measure at the link level, e.g., via SNMP \cite{fernandesComprehensiveSurveyNetwork2019}, as compared to collecting point-to-point flows performing packet inspection.
	However, if flows are directly available, the measurement reduces to $\bY = \bO \hprod(\bZ + \bA + \bN)$ with $E=F$, which corresponds to a choice of $\bR=\idmat$ in \eqref{eq:objfun_tens}.
	The model is further suitable for vehicular traffic flows \cite{sofuogluGLOSSTensorBasedAnomaly2022}, phasor measurements in power systems \cite{gaoIdentificationSuccessiveUnobservable2016}, wireless time-frequency power spectra \cite{dingRobustOnlineSpectrum2017, liRobustOnlinePrediction2022} and various image processing problems, such as image reconstruction or artifact removal \cite{miriyathanthrigeDeepUnfoldingIteratively2022}.

	\section{Low-Rank Tensor Recovery Algorithm}\label{sec:bsca_algs}
	To address \gls{ad} in network flows with classical optimization, we first review an anomaly recovery problem formulation based on \gls{rpca} in Sec.~\ref{sec:bsca_algs}-\ref{sec:svd_recovery}, before extending it to a low-rank tensor formulation in Sec.~\ref{sec:bsca_algs}-\ref{sec:cpd_recovery}.
	We propose to solve the latter with a \acrlong{bsca} optimization approach in Sec.~\ref{sec:bsca_algs}-\ref{sec:bsca_tensor}.
	To reduce the computational complexity, the tensor-based formulation is augmented and in turn solved in Sec.~\ref{sec:bsca_algs}-\ref{sec:relaxed_cpd_recovery} and \ref{sec:bsca_algs}-\ref{sec:bsca_tensor_rlx}, respectively.
	\subsection{Matrix-Factorization-Based Recovery Problem} \label{sec:svd_recovery}
	The recovery of $\bA$ by fitting the model in \eqref{eq:obs_model} to the observation $\bY$ requires the simultaneous estimation of the auxiliary variable $\bZ$.
	However, the resulting problem of minimizing $\frobnorm{\bO \odot \left(\bY - \bR \left( \bZ + \bA \right)\right)}^2$ \wrt $\bZ$ and $\bA$ is underdetermined.
	In \cite{mardaniDynamicAnomalographyTracking2013}, two assumptions are proposed to regularize the problem:
	\begin{enumerate}
		\item The normal flow matrix $\bZ$ exhibits low-rank characteristics. 
		In, e.g., network traffic, this property stems from the assumption that typical network usage follows patterns correlated between nodes and, thus, flows, inducing linear dependence among the rows of $\bZ$. 
		This characteristic is empirically verified in \cite{lakhinaStructuralAnalysisNetwork2004}.
		\item The anomaly matrix $\bA$ is sparse, containing predominantly zero-entries, as we expect anomalies to occur infrequently across both flows and time.
	\end{enumerate}
	Since $\bZ$ is merely assisting the recovery of $\bA$, the problem is simplified by directly focusing on $\bX= \bR\bZ$, which adopts the low-rank property of $\bZ$.
	The low-rank property of $\bX$ can be encouraged by a nuclear norm regularization term $\norm{\bX}_{\star}$, which is a tight convex relaxation of the rank of $\bX$.
	Since the nuclear norm is computationally costly to evaluate, the substitution $\bX= \bP \bQ^\T$ with $\bP \in \Rset^{E\times R}$ and $\bQ \in \Rset^{T\times R}$ for a specified rank $R$ is performed, and the nuclear norm regularization term is substituted by the minimization of $\frobnorm{\bP}^2 + \frobnorm{\bQ}^2$.
	Equivalence to the nuclear norm regularization applies if $R \geq \Rgt$, where $\Rgt$ is the rank of the ground truth flows  \cite{steffensCompactFormulationL212018}. %
	\par
	Sparsity is commonly enforced by minimizing the $\ell_1$-norm $\norm{\bA}_1$, which is a tight convex relaxation of the $\ell_0$-norm that counts nonzero entries.
	The resulting optimization problem for flow recovery, as proposed by \cite{mardaniDynamicAnomalographyTracking2013}, is given as
	\begin{align}
		\begin{split}
			\min_{\bP,\bQ, \bA} \quad & \frac{1}{2}\frobnorm{\bO \odot \left(\bY - \bP\bQ^\T - \bR\bA \right)}^2 \\[-1mm]
			& + \frac{\lambda}{2}\left(\frobnorm{\bP}^2 + \frobnorm{\bQ}^2\right) + \mu \norm{\bA}_1.\label{eq:orig_problem} 
		\end{split}
	\end{align}
	The regularization parameters $\lambda>0$ and $\mu>0$ balance the regularization terms against the model-fitting term.
	
	\subsection{CPD-Based Recovery Problem}\label{sec:cpd_recovery}
	The low-rank property extracts flow sequences or conversely time instances which are correlated across all nodes.
	However, it fails to impose structure on individual flow sequences.
	In practice, additional prior knowledge such as periodicity of flows in time, e.g., approximate daily repetition of normal network usage patterns, is available. 
	To further inject prior knowledge, the low-rank matrix model can be extended to a low-rank tensor model \cite{kasaiNetworkVolumeAnomaly2016, xieFastTensorFactorization2017, xuAnomalyDetectionRoad2019, wangAnomalyAwareNetworkTraffic2020, sofuogluGLOSSTensorBasedAnomaly2022, wangHankelstructuredTensorRobust2021, chenLowRankAutoregressiveTensor2022, liOrderpreservedTensorCompletion2022}.
	The data structure is subsequently imposed by tensor decomposition techniques such as the \gls{cpd} or the Tucker decomposition \cite{koldaTensorDecompositionsApplications2009, sidiropoulosTensorDecompositionSignal2017}.
	\par
	In this work, periodicity is captured by folding the size $T$ time dimension in model \eqref{eq:obs_model} into two tensor modes of dimension $T_1 \in \Nset$ and $T_2 \in \Nset$ as in \cite{xuAnomalyDetectionRoad2019}, where $T_1$ is the period and $T = T_1 T_2$.
	In particular, we define $\tO\in \{0, 1\}^{E \times T_1 \times T_2}$, $\tY\in \Rset^{E \times T_1 \times T_2}$ and $\tA\in \Rset^{F \times T_1 \times T_2}$ such that $\bO$, $\bY$ and $\bA$ is the mode-1 unfolding of $\tO$, $\tY$ and $\tA$, respectively.
	Taking $\bY$ as an example, we have $\matel{\tY}{j, t_1, t_2} = \matel{\bY}{j, t}$ where $t=t_1 + (t_2-1)T_1$, \ie, $t_1$ can be viewed as the fast-time and $t_2$ as the slow-time.
	\par
	Analogous to the matrix factorization $\bP\bQ^\T$, the normal data is expressed as a \gls{cpd} $\cpd{\bP, \bQ_1, \bQ_2}$ of maximum tensor rank $\cprank$ with factor matrices $\bP \in \Rset^{E \times \cprank}$, $\bQ_1 \in \Rset^{T_1 \times \cprank}$ and $\bQ_2 \in \Rset^{T_2 \times \cprank}$.
	The robust tensor-based recovery problem is subsequently formulated as
	\begin{align}
		&\min_{\bP, \bQ_1, \bQ_2, \tA}  \quad f_\mr{td}(\bP, \bQ_1, \bQ_2, \tA)\nonumber\\
		&\quad \text{where} \quad f_\mr{td}(\bP, \bQ_1, \bQ_2, \tA)=\label{eq:objfun_tens}\\
		&\mkern70mu \frac{1}{2}\frobnorm{\ttO \odot \left(\tY - \cpd{\bP, \bQ_1, \bQ_2} - \tA \tmprod{1} \bR \right)}^2\nonumber\\[-1mm]
		&\mkern70mu + \frac{\lambda}{2}\left(\frobnorm{\bP}^2 + \frobnorm{\bQ_1}^2 + \frobnorm{\bQ_2}^2\right) + \norm{\tM \hprod \tA}_1, \nonumber
	\end{align}
	where $\tM \in \Rset_{>0}^{F \times T_1 \times T_2}$ generalizes the scalar regularization $\mu$, and $\ttO = \tO \hprod \tW$ generalizes the selection $\tO$ with an arbitrary weight tensor $\tW \in \Rset_{>0}^{E \times T_1 \times T_2}$.
	Further note that $(\tA \tmprod{1} \bR)\unf{1} = \bR\bA$.
	Similar to the low-rank matrix objective in \eqref{eq:orig_problem}, the term $\frac{\lambda}{2}\left(\frobnorm{\bP}^2 + \frobnorm{\bQ_1}^2 + \frobnorm{\bQ_2}^2\right)$ penalizes the tensor rank of the model $\cpd{\bP, \bQ_1, \bQ_2}$. 
	In fact, it can be shown that this regularization term leads to the sparsification of groups of columns $(\matel{\bP}{:,r}, \matel{\bQ_1}{:,r}, \matel{\bQ_2}{:,r})$ \cite{bazerqueRankRegularizationBayesian2013}.
	\begin{rem}
		Various alternative tensor models to the one employed in \eqref{eq:objfun_tens} have been explored.
		While Hankelization of the time dimension has been studied in \cite{kasaiNetworkVolumeAnomaly2016, wangHankelstructuredTensorRobust2021} to incorporate additional structure, this approach has the problem of introducing redundancy \cite{yamamotoFastAlgorithmLowrank2022}. %
		In \cite{xieGraphBasedTensor2018, yuSLRTASparseLowrank2021, wangAnomalyAwareNetworkTraffic2020}, the authors fold the flows at a single time step into two modes, where traffic source nodes and sink nodes are captured by different modes.
		If multi-mode data is available, e.g., each flow is characterized by multiple features such as packet rate and byte rate, the feature vector can be captured by a tensor mode \cite{tanTrafficVolumeData2013, streitNetworkAnomalyDetection2020, zhaoBayesianRobustTensor2016}.
		A recovery problem formulation closely related to \eqref{eq:objfun_tens}, including missing observations and linear compression of flows, is proposed in \cite{kasaiNetworkVolumeAnomaly2016}, although the tensor model relies on Hankelization of the time dimension.
	\end{rem}
	\subsection{BSCA Algorithm} \label{sec:bsca_tensor}
	To tackle Problem \eqref{eq:objfun_tens}, we propose a block-iterative scheme based on the \gls{bsca} framework \cite{yangInexactBlockCoordinate2020} with block variables $\bP$, $\bQ_1$, $\bQ_2$ and $\tA$.
	Compared to classic \gls{bcd} \cite{bertsekasNonlinearProgramming1999}, the \gls{bsca} framework allows for a more flexible choice of block variables while still obtaining closed-form updates.
	\par
	The block variables update for $\bP$, $\bQ_1$ and $\bQ_2$ can be obtained in closed form as the local minimizer of $f_\mr{td}$ when fixing all other block variables, respectively.
	It follows
	\begin{subequations}
	\begin{align}
		\bP_{\mr{td}}(\bQ_1, \bQ_2, \tA) &= \argmin_{\bP} f_\mr{td}(\bP, \bQ_1, \bQ_2, \tA) \nonumber\\[-2mm]
		&\mkern-130mu= \Bigl[\left((\bQ_2 \krprod \bQ_1)^\T \Diag{\matel{\ttO\unf{1}^2}{j,:}} (\bQ_2 \krprod \bQ_1) +\lambda\idmat\right)^{-1}\nonumber\\
		&\mkern-130mu \quad~ (\bQ_2 \krprod \bQ_1)^\T \Diag{\matel{\ttO\unf{1}^2}{j,:}}\nonumber\\[-1mm]
		&\mkern-130mu\quad \left(\matel{\bY\unf{1}}{j,:} - \matel{(\tA \tmprod{1}\bR)\unf{1}}{j, :} \right)| j=1,\dots, E \Bigr]^\T\!\!, \label{eq:pupdate_tens}\displaybreak[1]\\
		\bQ_{1,\mr{td}}(\bP, \bQ_2, \tA) &= \argmin_{\bQ_1} f_\mr{td}(\bP, \bQ_1, \bQ_2, \tA) \nonumber\\[-2mm]
		&\mkern-130mu= \Bigl[\left((\bQ_2 \krprod \bP)^\T \Diag{\matel{\ttO\unf{2}^2}{t_1,:}} (\bQ_2 \krprod \bP) +\lambda\idmat\right)^{-1}\nonumber\\
		&\mkern-130mu \quad~ (\bQ_2 \krprod \bP)^\T \Diag{\matel{\ttO\unf{2}^2}{t_1,:}}\nonumber\\[-1mm]
		&\mkern-130mu\quad \left(\matel{\bY\unf{2}}{t_1,:} - \matel{(\tA \tmprod{1}\bR)\unf{2}}{t_1, :} \right)| t_1=1,\dots, T_1 \Bigr]^\T\!\!, \label{eq:q1update_tens}\displaybreak[1]\\
		\bQ_{2,\mr{td}}(\bP, \bQ_1, \tA) &= \argmin_{\bQ_2} f_\mr{td}(\bP, \bQ_1, \bQ_2, \tA) \nonumber\\[-2mm]
		&\mkern-130mu= \Bigl[\left((\bQ_1 \krprod \bP)^\T \Diag{\matel{\ttO\unf{3}^2}{t_2,:}} (\bQ_1 \krprod \bP) +\lambda\idmat\right)^{-1}\nonumber\\
		&\mkern-130mu \quad~ (\bQ_1 \krprod \bP)^\T \Diag{\matel{\ttO\unf{3}^2}{t_2,:}}\nonumber\\[-1mm]
		&\mkern-130mu \quad \left(\matel{\bY\unf{3}}{t_2,:} - \matel{(\tA \tmprod{1}\bR)\unf{3}}{t_2, :} \right)| t_2=1,\dots, T_2 \Bigr]^\T\!\! \label{eq:q2update_tens}.
	\end{align}
	\end{subequations}
	\par
	The local minimization of $f_\mr{td}$ \wrt $\tA$ does not have a closed-form.
	To obtain an update of $\tA$ in closed-form, $f_\mr{td}$ is first approximated around the current iterate $\tA_0$ of $\tA$ to
	\begin{align}
		\tilde{f}_\mr{td}(\tX, \tA, \tA_0) &= \frac{1}{2}\sum_{i=1}^{F}\left\lVert\ttO \!\odot\! \left(\tY \!- \!\tX\! -  \matel{\tA_0}{\neg \!i,:,:} \tmprod{1} \matel{\bR}{:,\neg \! i}\right.\right.\nonumber\\ 
		&\quad \left.\left. - \matel{\bR}{:,  i} \toprod \matel{\tA}{i,:,:}  \right)\right\rVert_\upm{F}^2 + \norm{\tM \hprod \tA}_1,\!
	\end{align}
	where $\tX = \cpd{\bP, \bQ_1, \bQ_2}$ (c.f. best-response approximation \#1 in \cite{yangInexactBlockCoordinate2020}).
	The approximation decouples the slices $\matel{\tA}{i,:,:}$ for $i=1, \dots, F$ such that the closed-form minimizer
	\begin{align}
		\widetilde{\tA}(\tX, \tA) &= \argmin_{\tA}\tilde{f}_\mr{td}(\tX, \tA, \tA_0) \nonumber\\
		&= \sfttresh{\tM}{\left[\ttO^2 \odot \left(\tY - \tX - \tA_0 \tmprod{1} \bR\right)\right] \tmprod{1} \bR^\T \right.\nonumber\\ 
			&\quad\left. + \left[ \ttO^2 \tmprod{1} \bR^{2\T}\right] \odot \tA_0 } \oslash \left( \ttO^2 \tmprod{1} \bR^{2\T}\right) \label{eq:da_update}
	\end{align}
	is attained.
	Unlike \gls{bcd}, where slices are optimized sequentially, the next iterate of $\tA$ in the proposed method is obtained by the joint update across all slices
	\begin{align}
		\tA' = \tA_0 + \gamma^* (\tX, \widetilde{\tA}, \tA_0) \big(\widetilde{\tA} - \tA_0 \big) \label{eq:a_update} %
	\end{align}
	where $\gamma^*$ is a step size. 
	\begin{figure*}
		\vspace{-3mm}
		\begin{align}
			\gamma^* (\tX, \widetilde{\tA}, \tA_0) &= \argmin_\gamma \gamma \left(\norm{\tM\hprod\widetilde{\tA}}_1  - \norm{\tM\hprod\tA_0}_1 \right) + \frac{1}{2}\frobnorm{\ttO \odot \left(\tY - \tX -\left[ \tA_0 + \gamma (\widetilde{\tA} - \tA_0) \right] \tmprod{1} \bR \right)}^2 \nonumber\\[-1mm]
			&= \left[- \frac{\matel{\ttO^2 \odot \left(\tX + \tA_0 \tmprod{1} \bR - \tY\right) \hprod \left( (\widetilde{\tA} - \tA_0)\right)\tmprod{1}\bR }{\Sigma} + \norm{\tM \hprod \widetilde{\tA}}_1 - \norm{\tM \hprod\tA_0}_1}{\frobnorm{\ttO \odot \left( (\widetilde{\tA} - \tA_0) \tmprod{1} \bR \right)}^2}\right]_0^1 \label{eq:opt_step_size}
		\end{align}
		\vspace{-3mm}
	\end{figure*}
	Similar to \cite{yangInexactBlockCoordinate2020}, we determine the step size in closed-form as the exact minimizer in $\gamma^*$ of a majorizing function of the objective $f_\mr{td}$ at the updated point $\tA + \gamma (\widetilde{\tA} - \tA)$ as given in \eqref{eq:opt_step_size}.
	Note that the minimized objective in \eqref{eq:opt_step_size} results from bounding the regularization as $\norm{\tA + \gamma (\widetilde{\tA} - \tA)}_1 \leq \abs{1-\gamma}\norm{\tA}_1 + \abs{\gamma} \norm{\widetilde{\tA}}_1$, and that $[\cdot]_0^1$ is the projection to the interval $[0, 1]$.
	\begin{algorithm}[t]
		\caption{{\tbsca}}\label{alg:tens_bsca}
		\begin{algorithmic}
			\State \textbf{input} $\bP\idx{0}, \bQ_1\idx{0}, \bQ_2\idx{0}, \bA\idx{0}, \lambda, \tM, \tW,L$
			\For{$\ell=1,\dots,L$}
			\State $\bP\idx{\ell} \gets \bP_{\mr{td}}(\bQ_1\idx{\ell-1}, \bQ_2\idx{\ell-1}, \tA\idx{\ell-1})$ (see \eqref{eq:pupdate_tens})
			\State $\bQ_1\idx{\ell} \gets \bQ_{1,\mr{td}}(\bP\idx{\ell}, \bQ_2\idx{\ell-1}, \tA\idx{\ell-1})$ (see \eqref{eq:q1update_tens})
			\State $\bQ_2\idx{\ell} \gets \bQ_{2,\mr{td}}(\bP\idx{\ell}, \bQ_1\idx{\ell}, \tA\idx{\ell-1})$ (see \eqref{eq:q2update_tens})
			\State $\tX\idx{\ell} \gets [\![\bP\idx{\ell}, \bQ_1\idx{\ell}, \bQ_2\idx{\ell}]\!]$
			\State $\widetilde{\tA}\idx{\ell} \gets \widetilde{\tA}(\tX\idx{\ell}, \tA\idx{\ell-1})$ (see \eqref{eq:da_update})
			\State $\gamma\idx{\ell} \gets \gamma^* (\tX\idx{\ell}, \widetilde{\tA}\idx{\ell}, \tA\idx{\ell-1})$ (see\eqref{eq:opt_step_size})
			\State $\tA\idx{\ell} = \bA\idx{\ell-1} + \gamma\idx{\ell} (\widetilde{\tA}\idx{\ell} - \tA\idx{\ell-1})$
			\EndFor
			\State \textbf{return} $\bP\idx{L}, \bQ_1\idx{L}, \bQ_2\idx{L}, \bA\idx{L}$
		\end{algorithmic}
	\end{algorithm}
	\par
	The \gls{bsca}-based optimization algorithm for problem \eqref{eq:orig_problem} is summarized in Alg.~\ref{alg:tens_bsca}.
	It is guaranteed to monotonically improve the objective in \eqref{eq:orig_problem} in each iteration.
	\begin{prop}
		Any limit point of the sequence $\oset{\bP\idx{\ell}, \bQ_1\idx{\ell}, \bQ_2\idx{\ell}, \bA\idx{\ell}}_\ell$ returned by Alg.~\ref{alg:tens_bsca} for $L\to\infty$ is a stationary point of \eqref{eq:objfun_tens}.
	\end{prop}
	This is a direct consequence of \cite[Thm. 2]{yangInexactBlockCoordinate2020} under the cyclic update rule, since any subsequence of convergent sequences converges to the same limit points.
	Alg.~\ref{alg:tens_bsca} generalizes \cite[Alg.~1]{schynolDeepUnrollingAnomaly2023}, \ie, it reduces to \cite[Alg.~1]{schynolDeepUnrollingAnomaly2023} if $T_2 = 1$ and $\bQ_2$ is fixed as $\onevec$. 
	Comparing the iteration complexity to that of the algorithm in \cite[Alg.~1]{mardaniDynamicAnomalographyTracking2013} which solves \eqref{eq:orig_problem}, we only need to consider the update of $\bA$.
	While \cite[Alg.~1]{mardaniDynamicAnomalographyTracking2013} requires $\ocplx{\max \uset{F^2T ,EFRT}}$ multiply-add operations, the proposed \gls{bsca}-based algorithm requires $\ocplx{\max \uset{E^2T,EFT,ERT}}$.
	Thus, under the reasonable assumptions $R \leq E,F$ and $E < F$, the proposed algorithm exhibits a reduced computational cost.
	\subsection{Augmented CPD-Based Recovery Problem}\label{sec:relaxed_cpd_recovery}
	Each update of $\bP$, $\bQ_1$ or $\bQ_2$ requires solving $E$, $T_1$ or $T_2$ linear systems of equations corresponding to the solutions provided in \eqref{eq:pupdate_tens}-\eqref{eq:q2update_tens}, with each system having a different left-hand side.
	This is caused by the element-wise multiplication with $\ttO$ and it is a major computational bottleneck in large scenarios.
	To mitigate the problem, we introduce the auxiliary optimization variable $\ttX \in \Rset^{E \times T_1 \times T_2}$ and augment the minimization problem to
	\begin{align}
		& \min_{\ttX, \bP, \bQ_1, \bQ_2, \tA} f_\mr{aug}(\ttX, \bP, \bQ_1, \bQ_2, \tA)\displaybreak[1]\label{eq:objfun_tensrlx}\\
		&\mkern32mu\text{where} \mkern25mu f_\mr{aug}(\ttX, \bP, \bQ_1, \bQ_2, \tA) =\nonumber\\
		& \mkern110mu \frac{1}{2}\frobnorm{\ttO \odot \left(\tY - \ttX - \tA \tmprod{1} \bR \right)}^2 \nonumber\displaybreak[1]\\
		&\mkern110mu+\frac{\nu}{2} \frobnorm{\ttX - \cpd{\bP, \bQ_1, \bQ_2}}^2 + \norm{\tM \hprod\tA}_1\nonumber\displaybreak[1]\\
		&\mkern110mu + \frac{\lambda}{2}\left(\frobnorm{\bP}^2 + \frobnorm{\bQ_1}^2 + \frobnorm{\bQ_2}^2\right).\nonumber
	\end{align}
	The parameter $\nu >0$ regulates the fidelity of $\ttX$ to the \gls{cpd} model of the normal data, analogous to an augmented Lagrangian formulation.
	A similar technique is applied in \cite{liuExtendedSuccessiveConvex2022, tillmannDOLPHInDictionaryLearning2016}, where the auxiliary variable approximates a dictionary-based representation.
	It can be readily seen that in the limit $\nu \to \infty$ the optimal point of Problem \eqref{eq:objfun_tensrlx} is identical to the one of Problem \eqref{eq:objfun_tens}.
	\subsection{Augmented BSCA Algorithm}\label{sec:bsca_tensor_rlx}
	Applying the \gls{bsca} framework to the augmented problem in \eqref{eq:objfun_tensrlx}, the block update of $\ttX$ is obtained as
	\begin{align}
		&\ttX (\bP, \bQ_1, \bQ_2, \tA)\nonumber\\
		&~~= \argmin_{\ttX} f_\mr{aug}(\ttX, \bP, \bQ_1, \bQ_2, \tA) \nonumber\\
		&~~ = \left(\ttO^2 \!\!\odot (\bY - \tA\tmprod{1} \bR)  + \nu \cpd{\bP, \bQ_1, \bQ_2} \right) \!\oslash\! \left(\ttO^2 + \nu\onevec \right)\!. \label{eq:xupdate_tensrlx}
	\end{align}
	The updates of the factor matrices $\bP$, $\bQ_1$ and $\bQ_2$ simplify to regularized alternating least squares updates:
	\begin{subequations}
		\begin{align}
			&\bP_{\mr{aug}}(\ttX, \bQ_1, \bQ_2) \nonumber\\
			&~~= \argmin_{\bP} f_\mr{aug}(\tX, \bP, \bQ_1, \bQ_2, \tA) \nonumber\\[-3mm]
			&~~= \ttX\unf{1} (\bQ_2 \krprod \bQ_1) \left((\bQ_1^T \bQ_1) \hprod (\bQ_2^T \bQ_2) +\frac{\lambda}{\nu}\idmat\! \right)^{\!-1}\!,\label{eq:pupdate_tensrlx}\displaybreak[1]\\
			&\bQ_{1, \mr{aug}}(\ttX, \bP, \bQ_2) \nonumber\\
			&~~= \argmin_{\bQ_1} f_\mr{aug}(\tX, \bP, \bQ_1, \bQ_2, \tA) \nonumber\\[-3mm]
			&~~= \ttX\unf{2} (\bQ_2 \krprod \bP) \left((\bP^T \bP) \hprod (\bQ_2^T \bQ_2) +\frac{\lambda}{\nu}\idmat \right)^{\!-1}\!, \label{eq:q1update_tensrlx}\displaybreak[1]\\
			&\bQ_{2, \mr{aug}}(\ttX, \bP, \bQ_1) \nonumber\\
			&~~= \argmin_{\bQ_2} f_\mr{aug}(\tX, \bP, \bQ_1, \bQ_2, \tA) \nonumber\\[-3mm]
			&~~= \ttX\unf{3} (\bQ_1 \krprod \bP) \left((\bP^T \bP) \hprod (\bQ_1^T \bQ_1) +\frac{\lambda}{\nu}\idmat \right)^{\!-1}\!.\label{eq:q2update_tensrlx}
		\end{align}
	\end{subequations}
	The block-update of $\tA$ remains identical to Alg.~\ref{alg:tens_bsca} of the non-augmented formulation.
	\par
	In view of the directional edge flows in a network, normal flows can additionally be assumed to be nonnegative, \ie, $\bX = \bR\bZ \geq \nullvec$ where ``$\geq$'' is to be understood as elementwise.
	Note that anomalies may still be negative.
	Integrating this assumption into the augmented problem in \eqref{eq:objfun_tensrlx} as 
	\begin{align}
		\begin{split}
			\min_{\ttX, \bP, \bQ_1, \bQ_2, \tA} & \quad f_\mr{aug}(\ttX, \bP, \bQ_1, \bQ_2, \tA)\\
			\text{s.t.} &\quad \ttX \geq \nullvec \label{eq:objfun_tensrlxnnf},
		\end{split}
	\end{align}
	we obtain, since the problem can be decomposed into the elements of $\tX$, the straightforward modified block update
	\begin{align}
	\begin{split}
		&\ttX (\bP, \bQ_1, \bQ_2, \tA)\\
		&\mkern45mu = \Big[\left(\ttO^2 \odot (\bY - \tA\tmprod{1} \bR) + \nu \cpd{\bP, \bQ_1, \bQ_2} \right) \\
		&\mkern45mu \quad\oslash \left(\ttO^2 + \nu\onevec \right)\Big]_0^{\infty}, \label{eq:xupdate_tensrlxnnf}
	\end{split}
	\end{align}
	where $[\cdot]_0^{\infty}$ is the elementwise projection onto $[0, \infty)$.
	Adding a similar constraint $\cpd{\bP, \bQ_1, \bQ_2} \geq \nullvec$ to the problem in \eqref{eq:objfun_tens} leads to a nonnegative least squares problem in each update of $\bP$, $\bQ_1$ and $\bQ_2$, which itself requires a computationally costly iterative optimization approach.
	In comparison, the projection in \eqref{eq:xupdate_tensrlxnnf} incurs a negligible computational cost.
	\begin{algorithm}[t]
		\caption{{\tbscarlx}}\label{alg:tens_bsca_rlx}
		\begin{algorithmic}
			\State \textbf{input} $\bP\idx{0}, \bQ_1\idx{0}, \bQ_1\idx{0}, \bA\idx{0}, \lambda, \tM, \tW, \nu$
			\For{$\ell=1,\dots,L$}
			\State $\ttX\idx{\ell} \!\gets \ttX(\bP\idx{\ell-1}\!, \bQ_1\idx{\ell-1}\!\!, \bQ_2\idx{\ell-1}\!\!\!, \tA\idx{\ell-1})$ (\eqref{eq:xupdate_tensrlx} or \eqref{eq:xupdate_tensrlxnnf})
			\State $\bP\idx{\ell} \gets \bP_{\mr{aug}}(\ttX\idx{\ell}, \bQ_1\idx{\ell-1}, \bQ_2\idx{\ell-1})$ (see \eqref{eq:pupdate_tensrlx})
			\State $\bQ_1\idx{\ell} \gets \bQ_{1,\mr{aug}}(\ttX\idx{\ell}, \bP\idx{\ell}, \bQ_2\idx{\ell-1}$ (see \eqref{eq:q1update_tensrlx})
			\State $\bQ_2\idx{\ell} \gets \bQ_{2,\mr{aug}}(\ttX\idx{\ell}, \bP\idx{\ell}, \bQ_1\idx{\ell};)$ (see \eqref{eq:q2update_tensrlx})
			\State $\ttX\idx{\ell} \gets \ttX(\bP\idx{\ell}, \bQ_1\idx{\ell}, \bQ_2\idx{\ell}, \tA\idx{\ell-1})$ (\eqref{eq:xupdate_tensrlx} or \eqref{eq:xupdate_tensrlxnnf})
			\State $\widetilde{\tA}\idx{\ell} \gets \widetilde{\tA}(\ttX\idx{\ell}, \tA\idx{\ell-1}; \mu)$ (see \eqref{eq:da_update})
			\State $\gamma\idx{\ell} \gets \gamma^* (\ttX\idx{\ell}, \widetilde{\tA}\idx{\ell}, \tA\idx{\ell-1})$ (see \eqref{eq:opt_step_size})
			\State $\tA\idx{\ell} = \bA\idx{\ell-1} + \gamma\idx{\ell} (\widetilde{\tA}\idx{\ell} - \tA\idx{\ell-1})$
			\EndFor
			\State \textbf{return} $\ttX\idx{L},\bP\idx{L}, \bQ_1\idx{L}, \bQ_2\idx{L}, \bA\idx{L}$
		\end{algorithmic}
	\end{algorithm}
	\par
 	The resulting block-iterative algorithm is summarized in Alg.~\ref{alg:tens_bsca_rlx}.
 	The block variable $\ttX$ is updated twice per iteration $\ell$ to couple the update of $\tA$ with the immediately preceding update of the factorization $(\bP, \bQ_1, \bQ_2)$ and vice versa.
 	\begin{prop}
 		Any limit point of the sequence $\oset{\tX\idx{\ell}, \bP\idx{\ell}, \bQ_1\idx{\ell}, \bQ_2\idx{\ell}, \bA\idx{\ell}}_\ell$ returned by Alg.~\ref{alg:tens_bsca_rlx} for $L\to\infty$ is a stationary point of \eqref{eq:objfun_tensrlx} in case of $\ttX(\cdot)$ as in \eqref{eq:xupdate_tensrlx}, or a stationary point of \eqref{eq:objfun_tensrlxnnf} in case of $\ttX(\cdot)$ as in \eqref{eq:xupdate_tensrlxnnf}.
 	\end{prop}
 	The convergence result of Alg.~\ref{alg:tens_bsca_rlx} is again a consequence of \cite[Theorem 2]{yangInexactBlockCoordinate2020}.
	Albeit the augmentation substantially reduces the per-iteration complexity, the resulting algorithm empirically requires a good initialization of $\bP$, $\bQ_1$ and $\bQ_2$.
	This is remedied by exchanging the first iteration of Alg.~\ref{alg:tens_bsca_rlx} with one iteration of Alg.~\ref{alg:tens_bsca}.
	
	\section{Unrolled CPD-Based Anomaly Detection}\label{sec:unroll_tbsca}
	In this section, we propose \gls{dn} architectures based on Alg.~\ref{alg:tens_bsca} and Alg.~\ref{alg:tens_bsca_rlx}.
	First, we briefly motivate and specify the well-known \gls{auc} metric for \gls{ad} performance in Sec.~\ref{sec:unroll_tbsca}-\ref{sec:problemformulation}.
	In Sec.~\ref{sec:unroll_tbsca}-\ref{sec:deep_unrolling}, we discuss shortcomings of the anomaly recovery problem \eqref{eq:orig_problem} and its extensions in Sec.~\ref{sec:unroll_tbsca}-\ref{sec:cpd_recovery} in combination with the \gls{auc} metric, and explain how deep unrolling addresses them.
	This culminates in our proposed model-aided \gls{dn} architectures and their training procedure in Sec.~\ref{sec:unroll_tbsca}-\ref{sec:dl_bsca_tensor} to \ref{sec:unroll_tbsca}-\ref{sec:training}.
	\subsection{Metric for AD Performance}\label{sec:problemformulation}
	Given a recovered anomaly matrix $\bA\idx{L}$ after $L$ algorithm iterations, let
	\begin{equation}
		\widehat{\bA} = \abs{\bA\idx{L}} \big/ \max_{i,t}\uset*{\abs*{\matel{\bA\idx{L}}{i, t}}}\label{eq:scorenorm}
	\end{equation}
	be the normalized anomaly scores that are attributed to the flows at all time instances to a realization $\ssample$.
	Note that ``$\abs{\cdot}$'' is applied elementwise. %
	The decision whether a flow $i$ at time $t$ is considered anomalous is typically made by a common threshold $[\widehat{\bA}]_{i,t} \gtrless \delta$.
	The threshold $\delta$ is a design choice which is dictated by the minimum probability of detection or maximum probability of false alarms that is acceptable for an application.
	To be independent of this choice, we consider an unbiased estimate of the \gls{auc} of the \gls{roc} as our evaluation metric \cite{hanleyMeaningUseArea1982}:
	\begin{align}
		\auc(\bA\idx{L}; \mathcal{S}) &=\left(\abs{\setAgt_1(\ssample)}\abs{\setAgt_0(\ssample)}\right)^{-1}\nonumber\\ 
		&\mkern-20mu \sum_{\substack{\forall(i_1, t_1)\\\in \setAgt_1(\ssample)}} \sum_{\substack{\forall(i_0, t_0)\\\in \setAgt_0(\ssample)}} \stepfun\left(\matel{\hbA}{i_1,t_1} \!\!\!\!\!- \matel{\hbA}{i_0,t_0}\right). \label{eq:auc}
	\end{align}
	Here, $\setAgt_1(\ssample)$ denotes the set of indices of the ground-truth anomalies in a realization $\ssample$, the set $\setAgt_0(\ssample)$ is its complement and $\stepfun(\cdot)$ the step function\footnote{In some \gls{auc} definitions, $a=0$ has zero contribution towards the area.}
	\begin{align}
		\stepfun(a) = \left(\operatorname{sgn}(a) + 1\right)/2
	\end{align}
	The \gls{auc} is the probability of a method assigning a higher score to a true anomaly than to a normal sample \cite{hanleyMeaningUseArea1982}.
	It conveniently allows the comparison of methods independent of the detection threshold $\delta$.
	\subsection{Deep Unrolling}\label{sec:deep_unrolling}
	\begin{figure}
		\centering
\tikzsetnextfilename{main_heatmap_bsca_base}
\begin{tikzpicture}
\begin{axis}[
	ylabel near ticks,
	width=5.25cm,
	height=5.25cm,
	colorbar, %
	colormap/hot, %
	xmin=-6.1, xmax=2.1,
	ymin=-6.1, ymax=2.1,
	xlabel = {$\log(\lambda)$},
	ylabel = {$\log(\mu)$},
	point meta min=0.5,
	point meta max=0.75,
	]
	\addplot+[matrix plot*, mesh/cols=33, mark=none, point meta=explicit] table [x index=0, meta index=2] {data/gridsearch_BSCAtensor_nrlx_100iter_r200_on_cpd_s_cv0_iter100.txt};
	\node[circle, inner sep=0.5mm, fill=black, label={[align=center, font=\footnotesize]{max. AUC\\$(-0.25, -1.5)$}}] at (axis cs: -0.25, -1.5) {};
\end{axis}
\end{tikzpicture}
		\caption{Average \gls{auc} over the regularization parameters $(\lambda, \mu)$ after 100 iterations of {\bsca} for the data set S1. Grid resolution: $0.25$.\label{fig:heatmap}}
	\end{figure}
	We focus on two particular shortcomings of the classical algorithms.
	First, the minimization in \eqref{eq:orig_problem} does not exactly relate to the maximization of \gls{ad} performance in terms of the \gls{auc}.
	As later observed in Fig.~\ref{fig:comp_classical} in Sec.~\ref{sec:simulations}, it is advantageous for the \gls{ad} task to terminate the signal recovery algorithms early rather than to let them converge to the stationary point of problem \eqref{eq:objfun_tens}, \eqref{eq:objfun_tensrlx} or \eqref{eq:objfun_tensrlxnnf}.
	Secondly, the regularization parameters have to be extracted from data since they are a priori unknown. %
	In particular, the authors of \cite{paffenrothRobustPCAAnomaly2018} emphasize the importance of the appropriate choice of the sparsity parameter $\mu$ when utilizing \gls{rpca} for \gls{ad}.
	Fig.~\ref{fig:heatmap} corroborates that a wide range of values of the \gls{auc} is achievable by Alg.~\ref{alg:tens_bsca} depending on the regularization parameters $\lambda$ and $\tM =\mu\onetens$.
	Moreover, Fig.~\ref{fig:heatmap} suggests that the problem is particularly sensitive in terms of the \gls{auc} in the vicinity of the optimal regularization parameter values. %
	\par
	Deep unrolling/unfolding offers a way forward \cite{gregorLearningFastApproximations2010,mongaAlgorithmUnrollingInterpretable2021}.
	The concept entails grounding the architecture of a \gls{dn} $\model(\cdot, \btheta)$ on the structure of an iterative algorithm based on a data model, in this case Alg.~\ref{alg:tens_bsca} or Alg.~\ref{alg:tens_bsca_rlx}.
	Each iteration $\ell$ is considered as the $\ell$th layer of a \gls{dn}.
	The algorithm is then truncated to $L$ layers, and the layers are modified such that they incorporate learnable parameters $\btheta\idx{\ell}$. 
	The parameters $\btheta = \oset{\btheta\idx{\ell}}_{\ell=1}^L$ are optimized end-to-end based on an empirical data loss $J(\btheta, \dataset)$ for a data set $\dataset$. %
	Since the data loss can be supervised and only needs to be differentiable, its choice offers more flexibility compared the original algorithm objective.
	\subsection{Non-Adaptive Unrolled BSCA-Based RPCA}\label{sec:dl_bsca_tensor}
	\begin{figure*}[t]
		\begin{minipage}{0.486\textwidth}
		\centering
		\newcommand\xone{-0.3}
\newcommand\xtwo{1.3}
\newcommand\xthree{3.1}
\newcommand\xfour{4.3}
\newcommand\ylayone{-0.5}
\newcommand\ylayonex{\ylayone+0.7}
\newcommand\ylayonea{\ylayone-0.6}
\newcommand\dPQ{0.3}

\tikzsetnextfilename{main_unrolled_algo}
\begin{tikzpicture}[
	]
	
	\node[Mb, label={\scriptsize Layer 1}] (m1) at (-0.5, 2) {$\modelly_\mr{td}$};
	
	\node[Mb, fill=lightBlue, label=\scriptsize{Layer 2}] (m2) at (1.25, 2) {$\modelly_\mr{aug}$};
	\node[Mb, label={\scriptsize Layer $L$}] (m4) at (4.0, 2) {$\modelly_\mr{aug}$};
	\node[txt] (mdot) at (2.5, 2) {\large$\cdots$};
	\node[txt] (i0) at (-2, 2) {$\ttX\itidx{0}$\\$\bP\itidx{0}$\\$\bQ_1\itidx{0}$\\$\bQ_2\itidx{0}$\\$\tA\itidx{0}$};
	\node[txt] (iL) at (5.5, 2) {$\ttX\itidx{L}$\\$\bP\itidx{L}$\\$\bQ_1\itidx{L}$\\$\bQ_2\itidx{L}$\\$\tA\itidx{L}$};
	
	\draw[->, lsty] ([yshift=0mm]i0) -- ([yshift=0mm]m1.west);
	\draw[->, lsty] ([yshift=0mm]m1.east) -- ([yshift=0mm]m2.west);
	\draw[-, lsty] ([yshift=0mm]m2.east) -- ([yshift=0mm]mdot.west);
	\draw[->, lsty] ([yshift=0mm]mdot.east) -- ([yshift=0mm]m4.west);
	\draw[->, lsty] ([yshift=0mm]m4.east) -- ([yshift=0mm]iL);

	\node[txt, midGray] (xr) at (\xone-1.6, \ylayonex) {$\ttX\itidx{1}$};
	\node[txt] (pqr) at (\xone-1.55, \ylayone) {$\bP\itidx{1}\!, \!\bQ_1\itidx{1}\!\!,$\\$ \bQ_2\itidx{1}$};
	\node[txt] (ar) at (\xone-1.6, \ylayonea) {$\tA\itidx{1}$};
	\node[txt, midGray] (xd) at (\xfour+1.2, \ylayonex) {$\ttX\itidx{2}$};
	\node[txt] (pqd) at (\xfour+1.15, \ylayone) {$\bP\itidx{2}\!, \!\bQ_1\itidx{2}\!\!,$\\$ \bQ_2\itidx{2}$};
	\node[txt] (ad) at (\xfour+1.2, \ylayonea) {$\tA\itidx{2}$};

	\node[nr] (x1) at (\xone, \ylayonex) {$\ttX$};
	\node[nr] (x2) at (\xthree, \ylayonex) {$\ttX$};
	\node[nr] (q1) at (\xtwo, \ylayone) {$\bQ_1$};
	\node[nr] (p1) at (q1.west) [anchor=east] {$\bP$};
	\node[nr] (q2) at (q1.east) [anchor=west] {$\bQ_2$};
	\node[nr] (a1) at (\xfour, \ylayonea) {$\gamma^* | \tA$};
	
	\node[bdot] (aspl) at (\xone-0.65+\offs, \ylayonea) {};
	\node[bdot] (aspl2) at (\xtwo+1.1+\offs, \ylayonea-\offs) {};
	\node[bdot] (pqspl) at (\xone-0.7, \ylayone) {};
	\node[bdot] (pqspl2) at (\xtwo+1.1, \ylayone) {};
	\node[bdot] (xspl) at (\xthree+0.4, \ylayonex) {};
	
	\draw[->, lsty] ([yshift=0mm]pqr) -- (pqspl) |- ([yshift=-\offs cm]p1.west);
	\draw[->, lsty] ([yshift=0mm]pqr) -- (pqspl) |- ([yshift=\offs cm]x1.west);
	\draw[->, lsty] ([yshift=0mm]ar) -- (aspl) |- ([yshift=-\offs cm]a1.west);
	\draw[->, lsty] ([yshift=0mm]ar) -- (aspl) |- ([yshift=-\offs cm]x1.west);
	\draw[->, lsty] ([yshift=0mm]x1.east) -- (\xone+0.4, \ylayonex) |- ([yshift=+\offs cm]p1.west);
	\draw[->, lsty] ([yshift=0mm]q2.east) -- (pqspl2) |- ([yshift=+\offs cm]x2.west);
	\draw[->, lsty] ([yshift=0mm]q2.east) -- (pqspl2) -- ([xshift=2mm]pqd.west);
	\draw[->, lsty] ([yshift=0mm]aspl2) |- ([yshift=-\offs cm]x2.west);
	\draw[->, lsty] ([yshift=0mm]x2.east) -- (xspl) |- ([yshift=+\offs cm]a1.west);
	
	\draw[->, lsty, midGray] ([yshift=0mm]xspl) -- ([yshift=0mm]xd);
	\draw[->, lsty] ([yshift=0mm]a1.east) -- ([yshift=0mm]ad);
	
	\node[circred] (lam1) at (\xtwo+\offs, \ylayone-1.5) {$\lambda\idx{2}$};
	\node[circred] (nu1) at (\xone, \ylayone-1.5) {$\nu\idx{2}$};
	\node[circred] (mu1) at (\xfour, \ylayone-1.5) {$\mu\idx{2}$};
	
	\draw[->, lsty, lightRed] ([yshift=0mm]lam1) -- ([xshift=\offs cm]q1.south);
	
	\node[bdot, lightRed] (nuspl) at (\xone, \ylayone-0.9) {};
	\node[bdot, lightRed] (nuspl2) at (\xtwo-\offs, \ylayone-0.9) {};
	\draw[->, lsty, lightRed] ([yshift=0mm]nu1) -- (nuspl) -- ([yshift=0mm]x1.south);
	\draw[->, lsty, lightRed] ([yshift=0mm]nu1) -- (nuspl) -| ([yshift=0mm]x2.south);
	\draw[->, lsty, lightRed] ([yshift=0mm]nu1) -- (nuspl) -| ([xshift=-\offs cm]q1.south);
	
	\draw[->, lsty, lightRed] ([yshift=0mm]mu1) -- ([yshift=0mm]a1.south);
	
	\node[txt, midGray] (hsr) at (\xone-1.6, \ylayonex+0.55) {$\tY,\tO,\tR$};
	\node[bdot, midGray] (hsspl1) at (\xone, \ylayonex+0.55) {};
	\node[bdot, midGray] (hsspl2) at (\xthree, \ylayonex+0.55) {};
	\draw[->, lsty, midGray] (hsr) -| ([yshift=0cm]x1.north);
	\draw[->, lsty, midGray] (hsr) -| ([yshift=0cm]x2.north);
	\draw[->, lsty, midGray] (hsr) -| ([yshift=0cm]a1.north);

	\begin{scope}[on background layer]
		\node[nl, fit=(aspl)(pqspl)(nuspl)(hsspl1)(p1)(a1)(x1)(x2)(lam1), inner sep=1mm] {};
	\end{scope}

\end{tikzpicture}
		\vspace{-2mm}
		\caption{Block diagram of the unrolled {\utbscarlx}, where each layer $\modelly_\mr{td|aug}$ consists of the block updates of Alg.~\ref{alg:tens_bsca} or Alg.~\ref{alg:tens_bsca_rlx}, respectively. The parameters $\nu\idx{\ell}$, $\lambda\idx{\ell}$ and $\mu\idx{\ell}$ are learnable.\\~\\ \label{fig:unrolled_algo}}
		\end{minipage}
		\hfill
		\begin{minipage}{0.486\textwidth}
			\centering
			\newcommand\auxone{1.0}
\newcommand\auxtwo{3.0}
\newcommand\auylayone{-0.0}
\newcommand\auylaytwo{\auylayone - 1.3}

\tikzsetnextfilename{main_adapt_unrolled_algo}
\begin{tikzpicture}[
	]
	
	\node[Mb, label={\scriptsize Layer 1}] (m1) at (0.5, 2) {$\modelly_\mr{ad,td}$};
	\node[Mb, fill=lightBlue, label=\scriptsize{Layer 2}] (m2) at (2.25, 2) {$\modelly_\mr{ad,aug}$};
	\node[Mb, label={\scriptsize Layer $L$}] (m4) at (5.0, 2) {$\modelly_\mr{ad,aug}$};
	\node[txt] (mdot) at (3.625, 2) {\large$\cdots$};
	\node[txt] (i0) at (-1, 2) {$\ttX\itidx{0}$\\$\bP\itidx{0}$\\$\bQ_1\itidx{0}$\\$\bQ_2\itidx{0}$\\$\tA\itidx{0}$};
	\node[txt] (iL) at (6.5, 2) {$\ttX\itidx{0}$\\$\bP\itidx{L}$\\$\bQ_1\itidx{L}$\\$\bQ_2\itidx{L}$\\$\tA\itidx{L}$};
	
	\draw[->, lsty] ([yshift=0mm]i0) -- ([yshift=0mm]m1.west);
	\draw[->, lsty] ([yshift=0mm]m1.east) -- ([yshift=0mm]m2.west);
	\draw[-, lsty] ([yshift=0mm]m2.east) -- ([yshift=0mm]mdot.west);
	\draw[->, lsty] ([yshift=0mm]mdot.east) -- ([yshift=0mm]m4.west);
	\draw[->, lsty] ([yshift=0mm]m4.east) -- ([yshift=0mm]iL);

	\node[Pb] (mi) at (\auxtwo+1, \auylayone) {$\modelly_\mr{aug}$};
	\node[Pb] (fft1) at (\auxone, \auylaytwo) {$\embedfuni{\tW}$};
	\node[Pb] (fpar1) at (\auxtwo, \auylaytwo) {$\parfuni{\tW}$};
	\node[Pb] (fft2) at (\auxone, \auylaytwo-0.8) {$\embedfuni{\tM}$};
	\node[Pb] (fpar2) at (\auxtwo, \auylaytwo-0.8) {$\parfuni{\tM}$};
	\node[bdot] (itspl) at (\auxone-1-\offs, \auylayone) {};
	\node[bdot] (itspl2) at (\auxone-1-\offs, \auylaytwo-\offs) {};
	
	\node[txt] (i1) at (\auxone-2.0, \auylayone) {\phantom{?}\\\phantom{?}\\\phantom{?}\\ $\ttX\itidx{1}$\\ $\bP\itidx{1}$\\ $\bQ_1\itidx{1}$\\ $\bQ_2\itidx{1}$\\ $\tA\itidx{1}$};
	\node[txt] (i2) at (\auxtwo+3.5, \auylayone) {\phantom{?}\\\phantom{?}\\\phantom{?}\\ $\ttX\itidx{2}$\\$\bP\itidx{2}$\\ $\bQ_1\itidx{2}$\\ $\bQ_2\itidx{2}$\\ $\tA\itidx{2}$};
	
	\draw[->, lsty] ([yshift=0mm]i1) -- (itspl) |- ([yshift=-\offs cm]fft1.west);
	\draw[->, lsty] ([yshift=0mm]itspl) |- ([yshift=-\offs cm]fft2.west);
	\draw[->, lsty] ([yshift=0mm]i1) -- ([yshift=0cm]mi);
	\draw[->, lsty] ([yshift=0mm]mi) -- ([yshift=0cm]i2);
	
	\draw[->, lsty] ([yshift=\offs cm]fft1.east) -- node[midway, above, font=\footnotesize] {$\featvec_{\tW}\itidx{2}$} ([yshift=\offs cm]fpar1.west);
	\draw[->, lsty] ([yshift=0mm]fft1.east) --  ([yshift=0 cm]fpar1.west);
	\draw[->, lsty] ([yshift=-\offs cm]fft1.east) -- ([yshift=-\offs cm]fpar1.west);
	
	\draw[->, lsty] ([yshift=\offs cm]fft2.east) -- node[midway, above, font=\footnotesize] {$\featvec_{\tM}\itidx{2}$} ([yshift=\offs cm]fpar2.west);
	\draw[->, lsty] ([yshift=0mm]fft2.east) -- ([yshift=0 cm]fpar2.west);
	\draw[->, lsty] ([yshift=-\offs cm]fft2.east) -- ([yshift=-\offs cm]fpar2.west);
	
	\draw[->, lsty] ([yshift=\offs cm]fpar1.east) -| ([xshift=-\offs cm]mi.south);
	\draw[->, lsty] ([yshift=\offs cm]fpar2.east) -| ([xshift=0 cm]mi.south);
	
	\node[txt, midGray] (hsr) at (\auxone-2.0, \auylayone+0.65) {$\tY\!, \tO\!, \bR$};
	\node[bdot, midGray] (hsspl) at (\auxone-1+\offs, \auylayone+0.65) {};
	\node[bdot, midGray] (hsspl2) at (\auxone-1+\offs, \auylaytwo+\offs) {};
	\draw[->, lsty, midGray] (hsr) -- (hsspl) |- ([yshift=+\offs cm]fft1.west);
	\draw[->, lsty, midGray] (hsr) -- (hsspl) |- ([yshift=+\offs cm]fft2.west);
	\draw[->, lsty, midGray] (hsr) -| ([yshift=-\offs cm]mi);
	
	\node[ellipse, draw, fill=lightRed, inner sep=0mm, align=center, font=\footnotesize, minimum width=7mm, minimum height=6mm] (th1) at (\auxtwo+2, \auylaytwo+0.8-\offs) {$\lambda\itidx{\ell}\!,\!\nu\itidx{\ell}$};
	\node[circred] (th2) at (\auxtwo+2, \auylaytwo-\offs) {$\btheta_{\tW}\itidx{\ell}$};
	\node[circred] (th3) at (\auxtwo+2, \auylaytwo-0.8-\offs) {$\btheta_{\tM}\itidx{\ell}$};
	\draw[->, lsty] (th1) -| ([xshift=+\offs cm]mi.south);
	\draw[->, lsty] (th2) -- ([yshift=-\offs cm]fpar1.east);
	\draw[->, lsty] (th3) -- ([yshift=-\offs cm]fpar2.east);
	
	\begin{scope}[on background layer]
		\node[nl, fit=(mi)(fpar1)(fft1)(fpar2)(fft2)(itspl)(th1), inner sep=2mm] {};
	\end{scope}
\end{tikzpicture}
			\vspace{-2mm}
			\caption{Block diagram of the adaptive unrolled algorithm {\autbsca}, where $\modelly_{\mr{aug}}$ is one layer of {\utbscarlx}, $\embedfun$ is a permutation invariant feature map, $\parfun$ is a learnable map for the parameters $\tW$ and $\tM$, respectively, and $\featvec_{\tW}\idx{\ell}$ and $\featvec_{\tM}\idx{\ell}$ are the embeddings per link or flow and time step, respectively. \label{fig:adapt_unrolled_algo}}
		\end{minipage}
	\end{figure*}
	Let $\obsfun(\ssample)=(\tY, \tO, \bR)$ be the observation of a realization $\ssample$ and consider for now $\tM = \mu\onetens$ and $\tW=\onetens$.
	Denote one full iteration of Alg.~\ref{alg:tens_bsca} or Alg.~\ref{alg:tens_bsca_rlx} as layer $\modelly_{\mr{td}}(\bP, \bQ_1, \bQ_2, \tA; \obsfun(\ssample), \btheta_{\mr{td}}\idx{\ell})$, where $\btheta_{\mr{td}}\idx{\ell} = \oset{\lambda\idx{\ell}, \mu\idx{\ell}}$,
	or layer $\modelly_{\mr{aug}}(\bP, \bQ_1, \bQ_2, \tA; \obsfun(\ssample), \btheta_{\mr{aug}}\idx{\ell})$, where $\btheta_{\mr{aug}}\idx{\ell} = \oset{\lambda\idx{\ell}, \mu\idx{\ell}, \nu\idx{\ell}}$, respectively.
	The parameters $\btheta_{\mr{td}}\idx{\ell}$ or $\btheta_{\mr{aug}}\idx{\ell}$ denote the trainable weights of layer $\ell$.
	The proposed unfolded \gls{dn} architecture, referred to as {\utbscarlx}, is
	\begin{align}
		\model_{\mr{aug}}(\ssample; \btheta)=\left(\modelly_{\mr{aug}}\idx{L}\circ \dots \circ \modelly_{\mr{aug}}\idx{2} \circ\modelly_{\mr{td}}\idx{1}\right)(\obsfun(\ssample);\btheta), \label{eq:unrolled_layers}
	\end{align}
	where $\modelly_{\mr{td|aug}}\idx{\ell}$ is shorthand for $\modelly_{\mr{td|aug}}(\,\cdot\,; \obsfun(\ssample), \btheta_{\mr{td|aug}}\idx{\ell})$, $\btheta = \oset{\btheta\idx{1}_{\mr{td}}, \btheta\idx{2}_{\mr{td|aug}}, \dots, \btheta\idx{L}_{\mr{td|aug}}}$ are the model parameters and $\mr{td|aug} \in \{\mr{td},\mr{aug}\}$.
	A model according to \eqref{eq:unrolled_layers} is visualized in Fig.~\ref{fig:unrolled_algo}.
	Note that the first layer is in both cases based on the non-augmented problem, \ie, $\modelly_{\mr{td}}$. %
	\subsection{Adaptation of Regularization Parameters}\label{sec:dl_bsca_tensor_adapt}
	Since the regularization parameters that maximize the \gls{ad} performance \wrt the \gls{auc} depend on the distribution of the traffic flows, it is beneficial to view the minimization problem from the perspective of statistical models.
	When interpreting the objective in \eqref{eq:orig_problem} as the negative log-likelihood of the measurement $\bY$, the model-fitting term $\frac{1}{2}\frobnorm{\bO \odot \left(\bY - \bP\bQ^\T - \bR\bA \right)}^2$ can be viewed as the negative log-likelihood of a Gaussian distributed model-error $\bN$ given $\bX=\bP\bQ^\T$ and $\bA$, where all elements have variance $1$.
	At the same time, the relaxed rank regularization $\frac{\lambda}{2}\left(\frobnorm{\bP}^2 + \frobnorm{\bQ}^2\right)$ can be viewed either as a negative Laplacian log-prior on the sum of singular values of $\bX$ or as an elementwise Gaussian prior with variance $\frac{1}{\lambda}$ on $\bP$ and $\bQ$.
	The sparsity regularization $\mu \norm{\bA}_1$ imposes a negative Laplacian log-prior on the elements in $\bA$.
	\par
	To account for the varying statistical properties across flows and time instances in real-world data, additional degrees of freedom can be incorporated into the objective function.
	For instance, the authors of \cite{mardaniEstimatingTrafficAnomaly2016} consider a problem similar to \eqref{eq:orig_problem} and propose a regularization based on general second-order statistics.
	They impose a Gaussian prior with spatial and temporal correlation matrices $\bC_{\bP}$ or $\bC_{\bQ}$ on the columns of $\bP$ or $\bQ$, respectively, via the regularization term $\frac{\lambda}{2} \left(\trace{\bP^\T \bC_{\bP}^{-1} \bP} + \trace{\bQ^\T \bC_{\bQ}^{-1} \bQ}\right)$.
	These correlation matrices are learned from training data and need to be relearned in case of domain changes.
	\par
	In this work, we increase the degrees of freedom of the statistical model with the previously introduced parameters $\ttO= \tO \hprod \tW$ and $\tM$ in \eqref{eq:objfun_tens} and \eqref{eq:objfun_tensrlx}.
	These extensions do not incur additional computational cost compared to the use of $\tO$ and $\mu$.
	The elements of the weighting tensor $\tW$ can be interpreted as the inverse standard deviation of the data-fitting error at each edge $j$ and each time instant $t$. This corresponds to flow components, which are not characterized by a low rank model, having different magnitudes across snapshots $j$ and $t$.
	As a result, it reflects the assumption that not each snapshot is equally well approximated by the low-rank model.
	Similarly, the tensor $ \tM$ individualizes the sparsity of each flow $i$ at each time instant $t$.
	In Sec.~\ref{sec:simulations}-\ref{sec:simulations_unrolled}\ref{sec:simulations_unrolled_rw}, the elements of $\tW$ and $\tM$ significantly differ after learning mappings of $\tW$ and $\tM$ on a real-world data distribution, substantiating the model. 
	Note that $\tW$ introduces a scalar ambiguity in the objective, which is addressed in Section \ref{sec:unrolled_adaptive_bsca}.
	\subsection{Online-Adaptive Unrolled Tensor-BSCA Architecture}\label{sec:unrolled_adaptive_bsca}
	Instead of learning $\tW$ and $\tM$ directly, we propose an adaptive \gls{dn} that learns a parametric representation map of $\tW$ and $\tM$, thereby enabling an online adaptation to the domain and statistics of the current scenario.
	To achieve this, we modify the architecture $\model_{\mr{aug}}$ in \eqref{eq:unrolled_layers} and Fig.~\ref{fig:unrolled_algo} to the architecture $\model_{\mr{ad,aug}}$ shown in Fig.~\ref{fig:adapt_unrolled_algo}, which we denote as {\autbsca}.
	Each adaptive layer $\modelly_{\mr{ad,aug}}$ consists, additionally to the original layers, of a composition of static feature embedding functions $\embedfuni{\tW|\tM}$ and parameter representation maps $\parfuni{\tW|\tM}$ with learnable weights $\btheta_{\tW|\tM}\idx{\ell}$: %
	\begin{align}
		\matel{\tW\idx{\ell}}{j,t_1,t_2} = \parfuni{\tW}\left(\cdot, \btheta_{\tW}\idx{\ell}\right) \circ \matel{\embedfuni{\tW}(\cdot)}{j, t_1, t_2, :},\\
		\matel{\tM\idx{\ell}}{i,t_1,t_2} = \parfuni{\tM}\left(\cdot, \btheta_{\tM}\idx{\ell}\right) \circ \matel{\embedfuni{\tM}(\cdot)}{i, t_1, t_2,:}.
	\end{align}
	The regularization parameters $\lambda\itidx{\ell}$ and $\nu\itidx{\ell}$ remain as directly learnable weights. %
	The embedding functions $\embedfuni{\tW|\tM}$ at layer $\ell$ encode the statistics of the scenario into tensors $\tH_{\tW|\tM}$ consisting of feature vectors $\featvec_{\tW|\tM} \in \Rset^{H_{\tW|\tM}}$ for every link load $j=1,\dots,E$ or every flow $i = 1, \dots, F$ at all time steps $t=1,\dots,T$ from the observation and current iterates.
	The maps $\parfuni{\tW|\tM}: \Rset^{H_{\tW|\tM}} \to \oset{0, \infty}$ compute a corresponding statistical model parameter for the unrolled iteration $\model_{\mr{td|aug}}$ corresponding to each feature vector $\featvec_{\tW|\tM}$.
	\par
	The embedded feature tensor $\tH_{\tW}\itidx{\ell} = \embedfuni{\tW}(\obsfun(\ssample), \tX\idx{\ell-1}, \tA\idx{\ell-1})$ is a 4D-tensor constructed of $H_{\tW}$ (hidden) features $\featvec_{\tW}$ for each link-and-time step $(j, t)$, 	where $t=t_1 + (t_2-1)T_1$:
	\begin{equation}
		\matel{\embedfuni{\tW}(\cdot)}{j, t_1, t_2,:} = \bmat*{\matel{\embedfuni{\tW}^d(\cdot)}{j, t_1, t_2}\big|d=1,\dots,H_{\tW}}^\T.
	\end{equation}
	The embedding functions $\embedfuni{\tW}^d(\cdot)$ are chosen from the predefined set $\mathcal{F}_{\mr{ft},\tW}$ in Tab.~\ref{tab:embeddings}. %
	In particular, \#1 and \#2 estimate the flow variance and model fit error variance, respectively, along tensor slices for each tensor mode, while \#3 counts the number of flows that are superimposed on a particular edge.
	To improve adaptation to changing orders of magnitudes of the flows, the range of values of the embedded features is compressed by a log-transform.
	\par
	Similarly, the embedded feature tensor $\tH_{\tM}\itidx{\ell} = \embedfuni{\tM}(\obsfun(\ssample), \tX\idx{\ell-1}, \tA\idx{\ell-1})$ is a 4D-tensor constructed of $H_{\tM}$ features for each flow-and-time step $(i, t)$:
	\begin{equation}
		\matel{\embedfuni{\tM}(\cdot)}{i, t_1, t_2,:} =\bmat*{\matel{\embedfuni{\tM}^d(\cdot)}{i, t_1, t_2}\big|d=1,\dots,H_{\tM}}^\T.
	\end{equation}
	The embedding functions $\embedfuni{\tM}^d(\cdot)$ are chosen from the predefined set $\mathcal{F}_{\mr{ft},\tM}$ in Tab.~\ref{tab:embeddings} as well.
	Here, \#1 and \#2 estimate anomaly amplitudes by finding the maximum across slices of the data-fitting error tensor which has been projected into the flow domain, whereas \#5 counts the number of observed links corresponding to a particular flow.
	Note that the embedding functions in Tab.~\ref{tab:embeddings} have been distilled from a larger set of functions by model comparison on the scenario SA (see Sec.~\ref{sec:simulations}-\ref{sec:datasets}) with one-sided t-tests using a significance level $0.05$.
	\begin{table*}
		\caption{Proposed sets $\mathcal{F}_{\mr{ft},\tW}$ and $\mathcal{F}_{\mr{ft},\tM}$ of embedding functions $\embedfuni{\tW}^d(\cdot)$ and $\embedfuni{\tM}^d(\cdot)$, respectively. The embedding functions are transformed by logarithmization $\log(\cdot + \epsilon)$. Further let $\tE = \left([\tO^2\hprod(\tY - \tX)] \tmprod{1} \bR^\T\right) \hdiv \left( \tO^2 \tmprod{1} \bR^{2\T}\right)$ (error projected onto flows) \label{tab:embeddings}, and let $\tA\unf{m}^{\mr{var}}$ and $\tE\unf{m}^{\mr{var}}$ with $\matel{\tA\unf{m}^{\mr{var}}}{n} = \emvar{\matel{\tA\unf{m}}{n,:}}$ and $\matel{\tE\unf{m}^{\mr{var}}}{n} = \emvar{\matel{\tE\unf{m}}{n,:}}$ , respectively.}
		\begin{tabularx}{\linewidth}{Xrrr}
			\toprule
			\multicolumn{4}{l}{Embed. fun. $\exp([\embedfuni{\tW}^d(\cdot)]_{j, t_1, t_2}) - \epsilon$} \\[-10pt]
			& 1st mode & 2nd mode & 3rd mode\\
			\midrule
			\#1 & $\emvar[\matel{\tO}{j,:,:}]{\matel{\tY}{j,:,:}}$ & $\emvar[\matel{\tO}{:,t_1,:}]{\matel{\tY}{:,t_1,:}}$ & $\emvar[\matel{\tO}{:,:,t_2}]{\matel{\tY}{:,:,t_2}}$\\[3pt]
			\#2 & $\emvar[\matel{\tO}{j,:,:}]{\matel{(\tY-\tX)}{j,:,:}}$ & $\emvar[\matel{\tO}{:,t_1,:}]{\matel{(\tY-\tX)}{:,t_1,:}}$ & $\emvar[\matel{\tO}{:,:,t_2}]{\matel{(\tY-\tX)}{:,:,t_2}}$\\[3pt]
			\#3 & $\sum_{i'}^{F} \matel{\bR}{j, i'}$ & & \\
			\midrule
\multicolumn{4}{l}{Embed. fun. $\exp([\embedfuni{\tM}^d(\cdot)]_{i, t_1, t_2}) - \epsilon$} \\[-10pt]
			& 1st mode & 2nd mode & 3rd mode\\
			\midrule
			\#1 & $\max_{n_2, n_3} \abs{\matel{\tE}{i, n_2, n_3}}$ & $\max_{n_1, n_3} \abs{\matel{\tE}{n_1,t_1,n_3}}$ & $\max_{n_1,n_2} \abs{\matel{\tE}{n_1,n_2,t_2}}$\\[3pt]
			\#2 & $\max\limits_{n_2, n_3} \abs*{\matel{\tE}{i, n_2, n_3}} /\! \sqrt{\matel{\tE\unf{2}^{\mr{var}}}{n_2} \matel{\tE\unf{3}^{\mr{var}}}{n_3}}$ & $\max\limits_{n_1, n_2} \abs*{\matel{\tE}{n_1,t_1,n_3}}/\! \sqrt{\matel{\tE\unf{1}^{\mr{var}}}{n_1} \matel{\tE\unf{3}^{\mr{var}}}{n_3}}$ & $\max\limits_{n_2,n_3} \abs*{\matel{\tE}{i,n_2,n_3}}/\! \sqrt{\matel{\tE\unf{1}^{\mr{var}}}{n_1} \matel{\tE\unf{2}^{\mr{var}}}{n_2}}$\\[3pt]
			
			\#3 & $\emvar[\matel{\tO}{i,:,:}]{\matel{\tA}{i,:,:}}$ & $\emvar[\matel{\tO}{:, t_1,:}]{\matel{\tA}{:, t_1,:}}$ & $\emvar[\matel{\tO}{:, :, t_2}]{\matel{\tA}{:, :,t_2}}$\\[3pt]
			\#4 & $\max\limits_{n_2, n_3} \abs*{\matel{\tA}{i, n_2, n_3}} /\! \sqrt{\matel{\tA\unf{2}^{\mr{var}}}{n_2} \matel{\tA\unf{3}^{\mr{var}}}{n_3}}$ & $\max\limits_{n_1, n_2} \abs*{\matel{\tA}{n_1,t_1,n_3}}/\! \sqrt{\matel{\tA\unf{1}^{\mr{var}}}{n_1} \matel{\tA\unf{3}^{\mr{var}}}{n_3}}$ & $\max\limits_{n_2,n_3} \abs*{\matel{\tA}{i,n_2,n_3}}/\! \sqrt{\matel{\tA\unf{1}^{\mr{var}}}{n_1} \matel{\tA\unf{2}^{\mr{var}}}{n_2}}$\\[3pt]
			\#5 & $\matel{\tO \tmprod{1}\bR^\T}{i, t_1, t_2}$ & & \\
			\bottomrule
		\end{tabularx}
	\end{table*}
	It is straightforward in this framework to introduce additional features if available, for instance, features identifying device properties of network nodes corresponding to flows $i$.
	\par
	It is critical for a good generalization that the feature embedding functions preserve invariances of the system \cite{battagliaRelationalInductiveBiases2018}.
	For example, the permutation of flows or links, resulting from, \eg, a relabeling of the nodes and the edges, should result in the corresponding permutation of features.
	Both embedding functions $\embedfuni{\tW}$ and $\embedfuni{\tM}$ are by design permutation equivariant \cite{battagliaRelationalInductiveBiases2018} in the fast-time $t_1$ and slow-time $t_2$, links $j$ and flows $i$, \ie, the model estimate $\bA\idx{L} = \tA\idx{\ell}\unf{1}$ does not change except for a corresponding permutation\footnote{In future work, permutation equivariance in time can be refined to, \eg, shift invariance in time, which enables the utilization of 1-dimensional convolutional \glspl{nn} \cite{lecunBackpropagationAppliedHandwritten1989} for embedding instead.}.
	\par
	The parameter representation maps $\parfuni{\tW}$ and $\parfuni{\tM}$ are composed of a body $\parfuni{\tW|\tM, \mr{b}}: \Rset^{H_{\tW|\tM}} \to \Rset$, \eg, shallow \glspl{mlp}, and a head $\parfuni{\tW|\tM, \mr{h}}: \Rset \to \oset{0, \infty}$ such that $\parfuni{\tW|\tM} = \parfuni{\tW|\tM, \mr{h}} \circ \parfuni{\tW|\tM, \mr{b}}$. 
	As the function head, we propose
	\begin{equation}
		\parfuni{\tW|\tM, \mr{h}} (x) = \eul^{C\tanh(x/C)}.
	\end{equation}
	The exponentiation ensures positive entries in $\tW$ and $\tM$, whereas $\tanh(\cdot)$ approximates a linear function around $0$ and bounds the image of $\parfuni{\tW|\tM}$ to $(\eul^{-C}, \eul^{C})$.
	The bound inhibits gradient explosion during training \cite[Sec.~8.2.4]{goodfellowDeepLearning2016} and further addresses the scalar ambiguity present due to the regularization parameters in the objectives \eqref{eq:objfun_tens} and \eqref{eq:objfun_tensrlx}.
	
	\subsection{Training}\label{sec:training}
	The model parameters $\btheta$ are obtained as the minimizer of an empirical risk $J(\btheta; \dataset)$ given a data set $\dataset$ sampled from the distribution of scenario realizations $p_{\ssample}(\ssample)$.
	Adopting the objective functions in \eqref{eq:orig_problem}, \eqref{eq:objfun_tens} or \eqref{eq:objfun_tensrlx} as loss functions is not optimal from the perspective of \gls{ad} performance.
	Similarly, any supervised reconstruction-based loss function that minimizes the $\ell_1$ or $\ell_2$ distance between the low-rank matrix $\bX = \bR\bZ$ and its estimate $\hbX=\bP\itidx{L}(\bQ\itidx{L})^\T$ or between the true anomalies $\bA$ and their estimates $\hbA=\bA\itidx{L}$ \cite{solomonDeepUnfoldedRobust2020, cohenDeepConvolutionalRobust2019, vanluongDeepUnfoldedReferenceBasedRPCA2021, maiDeepUnrolledLowRank2022, joukovskyInterpretableNeuralNetworks2024, dongDeepUnfoldedTensor2023, caiLearnedRobustPCA2021, miriyathanthrigeDeepUnfoldingIteratively2022} is not optimal \wrt the decision task of whether an anomaly is present or not.
	The unrolled \gls{rpca} approach in \cite{joukovskyInterpretableNeuralNetworks2024}, additionally leverages the supervised cross-entropy between a binary ground-truth outlier mask and an estimated outlier mask.
	However, cross-entropy optimization also does not generally correspond to optimizing the trade-off between the probability of detection and false alarm \cite{yanOptimizingClassifierPerformance2003}.
	\par
	In comparison, the \gls{auc} estimate in \eqref{eq:auc} directly quantifies the detection performance. %
	To make it suitable as a loss function despite its non-differentiability, the step function $\stepfun(a)$ is replaced by the logistic function $\logistic_\beta(a)= 1 / (1 + \eul^{-\beta a})$, obtaining the $\beta$-soft \gls{auc} estimate \cite{yanOptimizingClassifierPerformance2003, caldersEfficientAUCOptimization2007}
	\begin{align}
		\sauc_\beta(\bA\idx{L}; \mathcal{S}) &=\left(\abs{\setAgt_1(\ssample)}\abs{\setAgt_0(\ssample)}\right)^{-1}\nonumber\\ & \mkern-50mu \sum_{\substack{\forall(i_1, t_1)\\\in \setAgt_1(\ssample)}} ~\sum_{\substack{\forall(i_0, t_0)\\\in \setAgt_0(\ssample)}} \logistic_\beta\left(\matel{\hbA}{i_1,t_1} \!\!\!\!\! - \matel{\hbA}{i_0,t_0}\right), \label{eq:sauc}
	\end{align}
	where $\beta > 0 $ is a deformation parameter and $\hbA$ is defined as in \eqref{eq:scorenorm}.
	The $\beta$-soft \gls{auc} in \eqref{eq:sauc} is a differentiable approximation of the unbiased \gls{auc} estimate and approaches the unbiased \gls{auc} estimate for $\beta \to \infty$. %
	Denoting the estimate of the anomalies by the proposed unrolling-based \gls{dn} models as $\bA\idx{L}(\obsfun(\ssample);\btheta)$ given a set of trainable parameters $\btheta$ and observations $\obsfun(\ssample)$, we propose the empirical risk as the average $\beta$-soft \gls{auc} over realizations
	\begin{align}
		J_\beta(\btheta; \dataset) = -\frac{1}{\abs{\dataset}} \sum_{\forall\ssample \in \dataset} \sauc_\beta(\bA\idx{L}(\obsfun(\ssample);\btheta); \ssample) \label{eq:lossfun}.
	\end{align}
	To learn the model weights $\btheta$, homotopy optimization \cite{dunlavyHomotopyOptimizationMethods2005} is performed by setting $\beta= \beta_0$ until optimization step $i=i_0$, after which it gradually increase $\beta$ to $\beta_1$ at step $i_1$, thereby converging towards the unbiased \gls{auc} estimate in \eqref{eq:auc} in the process.
	In conjunction, mini-batch \gls{sgd} over $N_\mathrm{step}$ steps is applied to minimize the loss $J_\beta(\btheta; \dataset)$ up to approximate first-order stationarity.
	Homotopy optimization can also be considered as a \textit{soft-to-hard} annealing approach \cite{agustssonSofttoHardVectorQuantization2017}.
	\par
	\textbf{Discussion.} On the one hand, employing the \gls{auc} estimator as the loss function has several advantages.
	First, neither the sign nor the amplitude of the ground-truth anomalies is required in \eqref{eq:auc}, simplifying the collection of training data.
	Secondly, the function is robust against erroneously labeled data.
	Consider a scenario where only a subset $\setestAgt_1 \subseteq \setAgt_1$ of all ground-truth anomalies $\setAgt_1$ in a realization $\ssample$ is labeled as anomalous, whereas $\setestAgt_0 \supseteq \setAgt_0$ is the complementary set of flows labeled as non-anomalous, which is contaminated by the set of ground-truth anomalies $\setAul_1 = \setAgt_1 \backslash \setestAgt_1$ that are labeled as non-anomalous.
	Let $\auc(\bA; \setestAgt_1, \setestAgt_0)$ be the \gls{auc} estimate based on the available labels, and let $\auc(\bA; \setestAgt_1, \setAgt_0)$ be the AUC estimate \wrt only the correctly labeled ground-truth anomalies and all ground-truth non-anomalies.
	It follows that (see Appendix \ref{sec:apx_auc_label_err})
	\begin{equation}
		\abs{\auc(\bA; \setestAgt_1, \setestAgt_0) - \auc(\bA; \setestAgt_1, \setAgt_0)} \leq \frac{\abs{\setAul_1}}{\abs{\setestAgt_0}} \label{eq:apx_auc_err}.
	\end{equation}
	Eq. \eqref{eq:apx_auc_err} bounds the error due to the contamination of the set of  flows labeled as non-anomalies $\setestAgt_0$ by $\abs{\setAul_1} /\abs{\setAgt_0}$ with ground-truth anomalies, which is approximately the frequency of erroneously labeled ground-truth anomalies $\abs{\setAul_1} /(FT)$ if it is sufficiently small.
	\par
	On the other hand, the computational cost of the empirical risk $J_\beta$ scales with $\ocplx{F^2T^2}$, which may be prohibitive for large domains.
	To mitigate this drawback, $J_\beta$ can be further approximated.
	Assuming estimated flow amplitudes and scores are independent across flows and time, the \gls{auc} estimate approximates $\expect{\stepfun\left(\hat{a}_1 - \hat{a}_0\right)}$, where $\hat{a}_1$ and $\hat{a}_0$ are the random variables of the anomalous and non-anomalous scores \cite{hanleyMeaningUseArea1982}.
	Similarly, we have $\sauc_\beta(\bA\idx{L}(\obsfun(\ssample);\btheta); \ssample) \approx \expect{\logistic_\beta\left(\hat{a}_1 - \hat{a}_0\right)}$.
	It follows that the loss function can be approximated by sampling the terms  $\logistic_\beta\big([\hbA]_{f_1,t_1} - [\hbA]_{f_0,t_0}\big)$.
	\par
	Here, we apply a deterministic sampling strategy.
	Let $\uset{\setAgt_{1,k}}_{k=1}^{K_\mr{sub}}$ and $\uset{\setAgt_{0,k}}_{k=1}^{K_\mr{sub}}$ be partitions of $\setAgt_1$ and $\setAgt_0$, respectively.
	The subsampled $\beta$-soft \gls{auc} estimate is proposed as
	\begin{align}
		&\widetilde{\sauc}_\beta^{K_\mr{sub}}(\bA\idx{L}; \mathcal{S}) =\sum_{k=1}^{K_\mr{sub}} \frac{1}{K_\mr{sub}\abs{\setAgt_{1,k}(\ssample)}\abs{\setAgt_{0,k}(\ssample)}}\nonumber\\
		&\qquad\qquad \sum_{\substack{\forall(i_1, t_1)\\\in \setAgt_{1,k}(\ssample)}} ~\sum_{\substack{\forall(i_0, t_0)\\\in \setAgt_{0,k}(\ssample)}} \!\!\!\!\logistic_\beta\left(\matel{\hbA}{i_1,t_1} \!\!\!\!\!\!- \matel{\hbA}{i_0,t_0}\right) \!\label{eq:sampsauc}.
	\end{align}
	Partitioning ensures that each score $[\hbA]_{i,t}$ contributes to the loss function, and choosing subsets with similar cardinality ensures that the gradient \wrt any score $[\hbA]_{i,t}$ is the average of a sufficiently large number of terms.

	\section{Empirical Results}\label{sec:simulations}
	\begin{table}[t]
		\setlength{\tabcolsep}{5pt}
		\centering
		\caption{Compared methods. \label{tab:exp_algos}}
		\begin{tabularx}{\linewidth}{clcl}
			\toprule
			& \textbf{Methods} & \textbf{Sec./Ref.} & \textbf{Attributes}\\
			\midrule
			\multirow{4}{*}{\rotatebox[origin=c]{90}{\textbf{Matrix}}} & {\bbcd} & \cite{mardaniDynamicAnomalographyTracking2013} & iterative\\
			& {\aubsca} & \cite{schynolDeepUnrollingAnomaly2023} & unrolled, adaptive\\
			& {\umbscarlx} (prop.) & \ref{sec:unroll_tbsca}-\ref{sec:dl_bsca_tensor} & unrolled, relaxed\\
			& {\aumbscarlx} (prop.) & \ref{sec:unroll_tbsca}-\ref{sec:dl_bsca_tensor_adapt} & unrolled, adaptive, relaxed\\
			\midrule
			\multirow{6}{*}{\rotatebox[origin=c]{90}{\textbf{Tensor}}} & {\kasai} & \cite{kasaiNetworkVolumeAnomaly2016} & iterative, Hankelization\\
			& {\tbsca} (prop.) & \ref{sec:bsca_algs}-\ref{sec:bsca_tensor} & iterative \\
			& {\tbscarlx} (prop.) & \ref{sec:bsca_algs}-\ref{sec:bsca_tensor_rlx} & iterative, relaxed \\
			& {\utbscarlx} (prop.) & \ref{sec:unroll_tbsca}-\ref{sec:dl_bsca_tensor} & unrolled, relaxed\\
			& {\autbsca} (prop.) & \ref{sec:unroll_tbsca}-\ref{sec:dl_bsca_tensor_adapt}& unrolled, adaptive, relaxed \\
			\midrule
		\end{tabularx}
	\end{table}
	In this section, the performance of both the proposed iterative recovery algorithms and the unrolled \gls{dn} architectures, summarized in Tab.~\ref{tab:exp_algos}, are evaluated on synthetic and real-world data.
	For {\utbscarlx} and {\autbsca}, their reductions to matrix factorization {\umbscarlx} and {\aumbscarlx}, \ie, $T_2=1$, are additionally evaluated.
	\par
	We first delineate the simulation setup in Sections \ref{sec:simulations}-\ref{sec:refalg} through \ref{sec:simulations}-\ref{sec:trainval_setup} before presenting empirical results from Sec.~\ref{sec:simulations}-\ref{sec:sims_classical} onward.
	All simulations\footnote{To promote reproducible research, the code is publicly available at https://github.com/lsky96/unrolled-tbsca-ad} are implemented in Python with the PyTorch \cite{paszkePyTorchImperativeStyle2019} package.
	
	\subsection{Reference Methods}\label{sec:refalg}
	A fair comparison to other methods is challenging due to different assumptions on data and especially their generalization capabilities.
	Several classical algorithms based on low-rank decomposition do not consider anomalies or missing observations.
	Regarding deep learning, generic \gls{nn} architectures do not adapt to domain changes such as varying network topologies or sequence lengths.
	Graph \glspl{nn} can adapt to changing graph topologies, and recurrent graph \glspl{nn} as well as convolutional \glspl{nn} can adapt to variable sequences lengths.
	However, the accommodation of signal compression in these methods is unclear, \eg, the indirect observation of flows $\bZ + \bA$ at edges as $\bY= \bR (\bZ + \bA)$ due a linear compression matrix $\bR$.
	We are unaware of deep-learning-based approaches, including the model-aided \gls{rpca} approaches in \cite{solomonDeepUnfoldedRobust2020, cohenDeepConvolutionalRobust2019, vanluongDeepUnfoldedReferenceBasedRPCA2021, maiDeepUnrolledLowRank2022, joukovskyInterpretableNeuralNetworks2024, dongDeepUnfoldedTensor2023, caiLearnedRobustPCA2021, miriyathanthrigeDeepUnfoldingIteratively2022}, that address a model as discussed in \eqref{eq:obs_model}, including multiple simultaneous flows, outlier localization (in time and space, \ie, the flow), indirectly measured flows, and incomplete measurement matrices.
	We therefore focus on the comparison to the following methods:
	\begin{itemize}
		\item The matrix-factorization-based batch \gls{bcd} algorithm from \cite{mardaniDynamicAnomalographyTracking2013}, that is based on the optimization problem in \eqref{eq:orig_problem}, and which we denote as {\bbcd};
		\item the unrolling-based algorithm from our preliminary work in \cite{schynolDeepUnrollingAnomaly2023} based on matrix factorization, which we denote as {\aubsca}; %
		\item the Hankel-tensor-based online \gls{ad} algorithm from \cite{kasaiNetworkVolumeAnomaly2016}, which is based on an \gls{cpd}-based optimization problem similar to \eqref{eq:objfun_tens}, thereby using a Hankelization of the time dimension with additional Hankel regularization instead of time folding, and which we denote as {\kasai}.
	\end{itemize}
	\subsection{Data Sets} \label{sec:datasets}
	Two types of data are employed in the simulations.
	\begin{table}
		\caption{Parameters of the synthetic and real-world data sets.\label{tab:synthetic_param}}
		\setlength{\tabcolsep}{4.5pt}
		\begin{tabularx}{\linewidth}{Xrrrrrrr}
			\toprule
			\multicolumn{1}{X}{\textbf{Scen. Par.}} & $N$ & $E$ & $F$ & $T_1$ & $T_2$ & &\\
			\midrule
			S1 & $10$ & $30$ & $90$ & $20$ & $10$ & & \\
			S2 & $15$ & $60$ & $210$ & $30$ & $10$ & & \\
			SA & $10$ & $50$ & $90$ & $10$ & $10$ & & \\
			RW & $11$ & $30$ & $110$ & $96$ & $14$ & & \\
			\toprule
			\multicolumn{1}{X}{\textbf{Sampling Par.}} & $\obsprob$ & $\Rgt$ & $s_\mr{min}$ & $s_\mr{max}$ & $\anoamp$ & $\anoprob$ & $\noisestd^2$\\
			\midrule
			S1 & $0.9$ & $30$ & $1.0$ & $1.0$ & $1.0$ & $0.005$ & $0.01$ \\
			S2 & $0.9$ & $70$ & $0.25$ & $1.0$ & $0.8$ & $0.005$ & $0.04$ \\
			SA & $0.95$ & $40$ & $0.25$ & $1.0$ & $1.5$ & $0.005$ & $0.25$ \\
			RW & $0.95$ & / & /  & / & $0.5$ & $0.01$ & / \\
			\bottomrule
		\end{tabularx}
	\end{table}
	\par
	\textbf{Synthetic data.} We generate synthetic scenario realizations $\ssample$ consisting of $N$-node graphs, where for each $\ssample$  random topologies with $E$ directional links and the corresponding routing matrix $\bR\in\uset{0,1}^{E\times F}$ are obtained as in \cite[Sec.~6-A]{mardaniDynamicAnomalographyTracking2013}.
	For each graph, a normal flow $\bZ = \tZ\unf{1} \in \Rset^{F\times T}$, an anomaly matrix $\bA = \tA\unf{1}\in\Rset^{F\times T}$ and a noise matrix $\bN\in\Rset^{E\times T}$ is drawn.
	\par
	In particular, let $\tS = \bs_1 \circ \bs_2 \circ \bs_3 \in \Rset^{F\times T_1 \times T_2}$ be a scaling tensor that mimics varying statistics over flows and time, which also varies across realization, and where the entries of $\bs_1 \in \oset{0, \infty}^{E}$, $\bs_2 \in \oset{0, \infty}^{T_1}$ and $\bs_3 \in \oset{0, \infty}^{T_2}$ are independently and randomly sampled from $\mathcal{U}(s_\mr{min}, s_\mr{max})$ with $ 0 < s_\mr{min}< s_\mr{max}$.
	The normal flows $\bZ = (\tS \hprod \widetilde{\tZ})\unf{1}$ are generated by sampling low-rank tensors $\widetilde{\tZ} = \cpd{\widetilde{\bZ}_1, \widetilde{\bZ}_2, \widetilde{\bZ}_3}/\Rgt$ with ground-truth tensor rank $\Rgt$, where the elements of the factor matrices $\widetilde{\bZ}_1 \in \Rset^{F\times \Rgt}$, $\widetilde{\bZ}_2 \in \Rset^{T_1\times \Rgt}$ and $\widetilde{\bZ}_3 \in \Rset^{T_2\times \Rgt}$ are randomly drawn from the exponential distribution $\operatorname{Exp}(1)$.
	An additive noise matrix $\bN = \bR (\tS \hprod \widetilde{\tN})\unf{1}$ is obtained by independently sampling the elements of $\widetilde{\tN}$ from a zero-mean Gaussian distribution with variance $\noisestd^2>0$.
	The anomalies are generated by $\bA = \anoamp(\tS \hprod \widetilde{\tA})\unf{1}$, where $\anoamp > 0$ is the anomaly amplitude and the elements of $\widetilde{\tA}$ are independently sampled from the categorical distribution with categories $(-1, 0, 1)$ and probabilities $(\anoprob/2, 1-\anoprob, \anoprob/2)$, \ie, $\anoprob$ is the probability of an anomaly.
	The elements of $\tO$ are sampled from the categorical distribution with categories $(0, 1)$ with probabilities $(1-\obsprob, \obsprob)$, \ie, a particular link load is observed with probability $\obsprob$.
	\par
	We generate two synthetic data sets S1 and S2 according to Tab.~\ref{tab:synthetic_param} containing $250$ and $500$ scenarios, respectively.
	\par
	\textbf{Real-world data.} The Abilene data set contains measured traffic flows of a U.S. internet backbone network over $24$ weeks \cite{zhangNetworkAnomography2005, xieTrafficDatsetsAbilene2024}.
	The equivalent network graph contains $N=11$ nodes and $E=30$ directed links.
	The routing matrix $\bR$ is fixed.
	To take advantage of periodicity in time, each realization $\ssample$ consists of $2$ weeks of consecutive flows with each time step being equivalent to \qty{15}{\min} of traffic, which corresponds to $T_1=96$ and $T_2=14$.
	Since the ground-truth is not available, we generate ground-truth anomalies by sampling elements of $\widetilde{\bA}$ as in the case of the synthetic data, then scale the elements by $\matel{\bA}{i, :} = \anoamp \max_{t}\{\matel{\bZ}{i,t}\} [\widetilde{\bA}]_{i, :}$.
	Based on the discussion in the context of the inequality \eqref{eq:apx_auc_err} we argue that the \gls{auc} error due a small number of incorrectly labeled anomalies that might already be contained in the flows in the Abilene data set is negligible.
	Finally, a real-world data set RW with parameters given in Tab.~\ref{tab:synthetic_param} containing $12$ samples is generated.
	
	\subsection{Training And Validation}\label{sec:trainval_setup}
	The parameters $\btheta$ of the \gls{dn} models are optimized with AdamW optimization \cite{loshchilovDecoupledWeightDecay2019} over $N_\mr{step}=20000$ steps for synthetic data and $N_\mr{step}=24000$ for real-world data, where the weight decay decreases from $0.05$ to $0.01$ at step $i=14000$.
	Unless specified otherwise, the loss function is the subsampled $\beta$-soft \gls{auc} $\widetilde{\sauc}_\beta^{K_\mr{sub}}$ in \eqref{eq:sampsauc}, where the number of partitions is ${K_\mr{sub}}=16$.
	In the process, the soft-to-hard annealing parameter $\beta$ is increased from $\beta_0= 10$ at training step $i_0=5000$ to $\beta_1= 100$ at training step $i_1= 11000$.
	The minibatch size and step size $\eta$ depend on the data set. 
	We use a minibatch size of $10$ for synthetic data and $3$ for real-world data.
	For the synthetic data, the step size $\eta$ decreases from $0.01$ to $0.01\cdot 0.25^5$; the step size for real-world data is additionally halved.
	The regularization parameters $\mu$ and $\tM$ are biased to avoid an all-zero output during training.
	For {\autbsca}, we use simple affine functions with $H_{\tW|\tM}+1$ learnable weights as the parametric map $\parfuni{\tW|\tM, \mr{b}}$.
	The regularization parameters of non-deep-learning methods are obtained by Bayesian optimization with Matern Gaussian process kernel \cite{snoekPracticalBayesianOptimization2012, gardnerBayesianOptimizationInequality2014, headScikitoptimize2021} on the training data \wrt the estimated \gls{auc}.
	All matrix-factorization-based algorithms use $R=\min\uset{E, T}$ as their maximum rank, all tensor-based algorithms use $\cprank=\min\uset{ET_1, ET_2, T_1T_2}$, which is a loose upper bound of the tensor rank \cite{koldaTensorDecompositionsApplications2009}.
	\par
	To approximate the uncertainty, cross-validation is used.
	We perform $5$-fold cross-validation for the synthetic data with training-validation split of $200$-$50$ for $S1$ and $400$-$100$ for S2.
	For the real-world data, we perform $4$-fold cross validation with training-validation split of $9$-$3$.
	
	\subsection{Convergence of Tensor-Based BSCA Algorithms}\label{sec:sims_classical}
	\begin{figure}[t]
		\centering
\tikzsetnextfilename{main_comparison_classical}
\begin{tikzpicture}
\begin{groupplot}[group style={group name=algcomp, group size=1 by 1, horizontal sep=2cm, vertical sep=0.5cm,}, height=5.5cm, width=\linewidth, ylabel near ticks]
	\nextgroupplot[
	xmin=0, xmax=500,
	ymin = 0.5, ymax = .8,
	minor tick num = 0,
	ytick={0.5, 0.6, 0.7, 0.8, 0.9, 1.0},
	xlabel = {Iteration $\ell$},
	ylabel = {AUC},
	mark repeat  = 200,
	legend style={legend columns=2},
	cycle list name = markcycle,
	]
	\addplot+[red, solid] table[col sep = tab, x index = 0, y index = 5]{data/classical/bayopt_bsca_r30_on_cpd_s_iter_combined.txt}; %
	\addplot+[blue, solid] table[col sep = tab, x index = 0, y index = 5]{data/classical/bayopt_bbcd_r30_on_cpd_s_iter_combined.txt}; %
	\addplot+[purple, solid] table[col sep = tab, x index = 0, y index = 5]{data/classical/bayopt_bsca_tens_nrlx_r200_on_cpd_s_iter_combined.txt};
	\addplot+[green, solid] table[col sep = tab, x index = 0, y index = 5]{data/classical/bayopt_bsca_tens_rlx_it1nrlx_r200_on_cpd_s_iter_combined.txt}; %
	\addplot [name path=bscaL,draw=none] table[col sep = tab, x index = 0, y index = 7] {data/classical/bayopt_bsca_r30_on_cpd_s_iter_combined.txt};
	\addplot [name path=bscaU,draw=none] table[col sep = tab, x index = 0, y index = 8] {data/classical/bayopt_bsca_r30_on_cpd_s_iter_combined.txt};
	\addplot [fill=red!25] fill between[of=bscaU and bscaL];
	\addplot [name path=bbcdL,draw=none] table[col sep = tab, x index = 0, y index = 7] {data/classical/bayopt_bbcd_r30_on_cpd_s_iter_combined.txt};
	\addplot [name path=bbcdU,draw=none] table[col sep = tab, x index = 0, y index = 8] {data/classical/bayopt_bbcd_r30_on_cpd_s_iter_combined.txt};
	\addplot [fill=blue!25] fill between[of=bbcdU and bbcdL];
	\addplot [name path=tbscanrlxL,draw=none] table[col sep = tab, x index = 0, y index = 7] {data/classical/bayopt_bsca_tens_nrlx_r200_on_cpd_s_iter_combined.txt};
	\addplot [name path=tbscanrlxU,draw=none] table[col sep = tab, x index = 0, y index = 8] {data/classical/bayopt_bsca_tens_nrlx_r200_on_cpd_s_iter_combined.txt};
	\addplot [fill=purple!25] fill between[of=tbscanrlxU and tbscanrlxL];
	\addplot [name path=tbscarlxitL,draw=none] table[col sep = tab, x index = 0, y index = 7] {data/classical/bayopt_bsca_tens_rlx_it1nrlx_r200_on_cpd_s_iter_combined.txt};
	\addplot [name path=tbscarlxitU,draw=none] table[col sep = tab, x index = 0, y index = 8] {data/classical/bayopt_bsca_tens_rlx_it1nrlx_r200_on_cpd_s_iter_combined.txt};
	\addplot [fill=green!25] fill between[of=tbscarlxitU and tbscarlxitL];
\legend{{\bsca}, {\bbcd},{\tbsca} (prop.),{\tbscarlx} (prop.)}
\end{groupplot}
\begin{axis}[
	at={($(algcomp c1r1.south east)+(-3mm,+4.5mm)$)},anchor=south east, axis background/.style={fill=gray!10},
	width=6.5cm,
	height=2.9cm,
	xmin=0, xmax=20,
	ymin = 0.5, ymax = 0.8,
	minor tick num = 0,
	mark repeat  = 4,
	legend style={legend columns=3},
	cycle list name = markcycle,
	]
	\addplot+[red, solid] table[col sep = tab, x index = 0, y index = 5]{data/classical/bayopt_bsca_r30_on_cpd_s_iter_combined.txt}; %
	\addplot+[blue, solid] table[col sep = tab, x index = 0, y index = 5]{data/classical/bayopt_bbcd_r30_on_cpd_s_iter_combined.txt}; %
	\addplot+[purple, solid] table[col sep = tab, x index = 0, y index = 5]{data/classical/bayopt_bsca_tens_nrlx_r200_on_cpd_s_iter_combined.txt};
	\addplot+[green, solid] table[col sep = tab, x index = 0, y index = 5]{data/classical/bayopt_bsca_tens_rlx_it1nrlx_r200_on_cpd_s_iter_combined.txt}; %
	\addplot [name path=bscaL,draw=none] table[col sep = tab, x index = 0, y index = 7] {data/classical/bayopt_bsca_r30_on_cpd_s_iter_combined.txt};
	\addplot [name path=bscaU,draw=none] table[col sep = tab, x index = 0, y index = 8] {data/classical/bayopt_bsca_r30_on_cpd_s_iter_combined.txt};
	\addplot [fill=red!25] fill between[of=bscaU and bscaL];
	\addplot [name path=bbcdL,draw=none] table[col sep = tab, x index = 0, y index = 7] {data/classical/bayopt_bbcd_r30_on_cpd_s_iter_combined.txt};
	\addplot [name path=bbcdU,draw=none] table[col sep = tab, x index = 0, y index = 8] {data/classical/bayopt_bbcd_r30_on_cpd_s_iter_combined.txt};
	\addplot [fill=blue!25] fill between[of=bbcdU and bbcdL];
	\addplot [name path=tbscanrlxL,draw=none] table[col sep = tab, x index = 0, y index = 7] {data/classical/bayopt_bsca_tens_nrlx_r200_on_cpd_s_iter_combined.txt};
	\addplot [name path=tbscanrlxU,draw=none] table[col sep = tab, x index = 0, y index = 8] {data/classical/bayopt_bsca_tens_nrlx_r200_on_cpd_s_iter_combined.txt};
	\addplot [fill=purple!25] fill between[of=tbscanrlxU and tbscanrlxL];
	\addplot [name path=tbscarlxitL,draw=none] table[col sep = tab, x index = 0, y index = 7] {data/classical/bayopt_bsca_tens_rlx_it1nrlx_r200_on_cpd_s_iter_combined.txt};
	\addplot [name path=tbscarlxitU,draw=none] table[col sep = tab, x index = 0, y index = 8] {data/classical/bayopt_bsca_tens_rlx_it1nrlx_r200_on_cpd_s_iter_combined.txt};
	\addplot [fill=green!25] fill between[of=tbscarlxitU and tbscarlxitL];
\end{axis}
\end{tikzpicture}
		\caption{Estimated validation \gls{auc} of non-unrolled methods averaged over 50 scenarios.\label{fig:comp_classical}}
	\end{figure}
	We first investigate the convergence and the corresponding achieved \gls{auc} of the proposed {\tbsca} of Sec.~\ref{sec:bsca_algs}-\ref{sec:bsca_tensor} and {\tbscarlx} of Sec.~\ref{sec:bsca_algs}-\ref{sec:bsca_tensor_rlx} against {\bbcd} \cite{mardaniDynamicAnomalographyTracking2013} and {\bsca} \cite{schynolDeepUnrollingAnomaly2023} on the data set S1.
	The regularization parameters are approximately optimally chosen in the sense that the \gls{auc} estimate is maximized within the first $100$ iterates.
	\par 
	Fig.~\ref{fig:comp_classical} compares the estimated \gls{auc} over the number of algorithm iterations.
	Overall, the tensor-based methods outperform {\bbcd} and {\bsca} in terms of the maximum achieved \gls{auc}, demonstrating their ability to effectively leverage the low-rank tensor structure of the data.
	{\tbscarlx} outperforms {\tbsca} since the relaxed objective in \eqref{eq:objfun_tensrlxnnf} offers an additional degree of freedom.
	However, despite the fact that the optimization objective is monotonically decreasing for all methods, the estimated \gls{auc} achieves a maximum at an intermediate iteration before decreasing again, supporting the claim that \gls{ad} performance in terms of the \gls{auc} and flow recovery based on model fitting and sparsity are only related objectives. %
	This motivates early stopping of the algorithms and tuning based on the task objective itself, which is accomplished by unrolling a fixed number of iterations.

	\subsection{Unrolled Tensor-Based BSCA Architectures}\label{sec:simulations_unrolled}
	\subsubsection{Validation of Loss Function}
	\begin{figure}[t]
		\centering
\tikzsetnextfilename{main_loss_auc_over_steps}
\begin{tikzpicture}
\begin{axis}[
	axis y line*=left,
	xmin=0, xmax=20000,
	ymin = 0.625, ymax = 0.75,
	width = .83\linewidth,
	height = 4.5cm,
	minor tick num = 0,
	xlabel = {Training Step},
	ylabel = {Neg. Loss $-J_\beta$},
	ylabel near ticks,
	legend style={legend columns=2},
	cycle list name = markcycle,
	]
	\addplot+[red, solid, mark repeat=200, mark phase=100] table[col sep = tab, x index = 0, y expr = {-\thisrowno{1}}]{data/synthetic/BSCATensorUnrolled_cpd_h_ly8_r300_mw_dff1lymuf1ly_rlx2x+nnf_it1nrlx_aucfs_tloss_combined.txt}; %
	\addplot+[blue, solid, each nth point=1, mark=triangle] coordinates{
		(0,1)
		(1,1) 
	}; %
	\addplot+[name path=trainL, draw=none, mark=none] table[col sep = tab, x index = 0, y expr = {-\thisrowno{4}}]{data/synthetic/BSCATensorUnrolled_cpd_h_ly8_r300_mw_dff1lymuf1ly_rlx2x+nnf_it1nrlx_aucfs_tloss_combined.txt}; %
	\addplot+[name path=trainU, draw=none, mark=none] table[col sep = tab, x index = 0, y expr = {-\thisrowno{3}}]{data/synthetic/BSCATensorUnrolled_cpd_h_ly8_r300_mw_dff1lymuf1ly_rlx2x+nnf_it1nrlx_aucfs_tloss_combined.txt}; %
	\addplot [fill=red, opacity=0.2] fill between[of=trainU and trainL];
	\node (arrow1) at (axis cs: 5000, 0.65) {$i_0, \beta=\beta_0$};
	\draw[-stealth, ultra thick] (arrow1)--($(arrow1)+(0.0cm,0.75cm)$);
	\node (arrow2) at (axis cs: 11000, 0.69) {$i_1, \beta=\beta_1$};
	\draw[-stealth, ultra thick] (arrow2)--($(arrow2)+(0.0cm,0.75cm)$);
	\legend{{Training Loss},{Validation AUC}}
\end{axis}
\begin{axis}[
	axis y line*=right,
	axis x line=none,
	xmin=0, xmax=20000,
	ymin = 0.625, ymax = 0.75,
	width = .83\linewidth,
	height = 4.5cm,
	minor tick num = 0,
	ylabel = {AUC},
	ylabel near ticks, %
	legend style={legend columns=2},
	cycle list name = markcycle,
	]
	\addplot+[blue, solid, each nth point=1, mark=triangle, mark repeat=100, mark phase=50] table[col sep = tab, x index = 0, y index = 5]{data/synthetic/BSCATensorUnrolled_cpd_h_ly8_r300_mw_dff1lymuf1ly_rlx2x+nnf_it1nrlx_aucfs_tepochs_combined.txt}; %
	\addplot+[name path=aucL, draw=none, each nth point=1, mark=none] table[col sep = tab, x index = 0, y index = 7]{data/synthetic/BSCATensorUnrolled_cpd_h_ly8_r300_mw_dff1lymuf1ly_rlx2x+nnf_it1nrlx_aucfs_tepochs_combined.txt}; %
	\addplot+[name path=aucU, draw=none, each nth point=1, mark=none] table[col sep = tab, x index = 0, y index = 8]{data/synthetic/BSCATensorUnrolled_cpd_h_ly8_r300_mw_dff1lymuf1ly_rlx2x+nnf_it1nrlx_aucfs_tepochs_combined.txt}; %
	\addplot [fill=blue, opacity=0.2] fill between[of=aucU and aucL];
\end{axis}
\end{tikzpicture}
		\caption{Training data loss and validation \gls{auc} estimate data loss over training steps of a {\autbsca} model with $L=8$ layers.\label{fig:tloss}}
	\end{figure}
	\begin{table}[t]
		\centering
		\caption{Estimated validation \gls{auc} achieved on the data set S2 of {\autbsca} over the number of partitions $K_\upm{sub}$ of the subsampled loss $\widetilde{\sauc}_\beta^{K_\mr{sub}}$ (left)  and the training set size $\abs{\dataset_\mr{train}}$ (right).\label{tab:lossfun_ss_trainsetsize}}
		\begin{tabularx}{.45\linewidth}{Xr}
			\toprule
			$K_\upm{sub}$ & \gls{auc} / $10^{-3}$\\
			\midrule
			1 & $725.5\pm0.6$\\
			2 & $725.1\pm1.0$\\
			4 & $724.3\pm1.6$\\
			8 & $724.9\pm0.7$\\
			16 & $725.0\pm0.9$\\
			32 & $724.0\pm0.4$\\
			\bottomrule
		\end{tabularx}
		\qquad
		\begin{tabularx}{.45\linewidth}{Xr}
			\toprule
			$\abs{\dataset_\mr{train}}$ & \gls{auc} / $10^{-3}$\\
			\midrule
			10 & $712.8\pm5.6$ \\
			25 & $722.3\pm2.0$ \\
			50 & $723.1\pm1.4$ \\
			100 & $724.8\pm1.0$ \\
			200 & $724.8\pm1.6$ \\
			400 & $725.0\pm0.9$ \\
			\bottomrule
		\end{tabularx}
	\end{table}
	In Tab.~\ref{tab:lossfun_ss_trainsetsize}, the subsampled $\beta$-soft \gls{auc} loss function is validated with the {\autbsca} architecture with $L=8$ layers.
	Partitioning scales the computational cost of the loss function in training by $1/K_\mr{sub}$.
	Increasing $K_\mr{sub}$ does not adversely affect the final estimated \gls{auc}, making it an effective strategy for reducing the computational complexity of the loss function.
	Note that the number of partitions $K_\mr{sub}$ is practically limited by the number of ground-truth anomalies in a data sample $\ssample$, \ie, $K_\mr{sub}$ cannot exceed it.
	We select $K_\mr{sub}=16$ for the remaining experiments.
	\par
	Fig.~\ref{fig:tloss} shows the training loss and \gls{auc} achieved on the validation data throughout the learning process.
	The negative loss yields an approximately increasing \gls{auc} estimate on the validation set over the training steps. 
	Indeed when $\beta$ is high, the loss function closely approximates the \gls{auc}.
	\subsubsection{Number of Layers}
	\begin{table*}[t]
		\setlength{\tabcolsep}{4pt}
		\centering
		\caption{Estimated validation AUC / $10^{-3}$ on the data set S2 over the number of model layers $\ell$. \label{tab:auc_layers}} 
		\begin{tabularx}{\linewidth}{Xrrrrrrrrr}
			\toprule
			Layer $\ell$ & 3 & 4 & 5 & 6 & 7 & 8 & 9 & 10 & 11\\
			\midrule
			{\aumbscarlx} & $694.2\!\pm\!3.4$ & $702.9\!\pm\!1.9$ & $705.3\!\pm\!2.2$ & $707.7\!\pm\!2.1$ & $709.3\!\pm\!2.4$ & $709.4\!\pm\!2.6$ & $711.1\!\pm\!2.1$ & $\mathbf{714.1\!\pm\!2.0}$ & $713.4\!\pm\!1.6$\\ %
			{\autbsca} & $705.9\!\pm\!0.9$ & $717.7\!\pm\!0.6$ & $717.8\!\pm\!2.4$ & $722.8\!\pm\!1.0$ & $723.3\!\pm\!1.7$ & $\mathbf{725.0\!\pm\!0.9}$ & $722.8\!\pm\!3.2$ & $-$& $-$\\ %
			\bottomrule
		\end{tabularx}
	\end{table*}
	Tab.~\ref{tab:auc_layers} compares the \gls{ad} performance over the number of model layers $L$.
	After rising with model depth, the validation \gls{auc} saturates for both the matrix-factorization-based and tensor-based architectures after a particular model depth which differs between architectures.
	We select in the following $L=8$ for the proposed tensor-based method and $L=10$ for the matrix-based method.

	\subsubsection{Training Data Efficiency}
	To evaluate the training data efficiency of our proposed method, we perform training on training data sets with varying cardinalities $\abs{\dataset_\mr{train}}$.
	We consider {\autbsca} as it has the highest number of learnable weights per layer among the proposed architectures.
	\par
	Tab.~\ref{tab:lossfun_ss_trainsetsize} demonstrates that while the best validation \gls{auc} is achieved for the largest training data set containing $400$ samples, the model remarkably achieves approx\onedot $98\%$ of the score when trained with only 10 samples.
	This efficiency is enabled by the architecture retaining the permutation invariances of the problem and having a low number of parameters --- {\autbsca} has only $24L-1$ learnable parameters --- compared to generic \gls{nn} architectures with easily thousands or even millions of parameters.

	\subsubsection{Comparison on Synthetic Data}
	\begin{table}[t]
		\centering
		\caption{Comparison of the estimated validation \gls{auc} on data set S2 achieved by different methods and configurations. \label{tab:comp_synthetic}}
		\begin{tabularx}{\linewidth}{Xrr}
			\toprule
			\textbf{Method} & $L$ & AUC for S2\\
			\midrule
			\multicolumn{3}{l}{\textbf{Reference Methods}}\\
			\midrule
			{\bbcd} \cite{mardaniDynamicAnomalographyTracking2013} & / & $681.0\pm0.9$ \\
			{\aubsca} \cite{schynolDeepUnrollingAnomaly2023} & 7 & $685.8\pm1.0$ \\
			{\kasai} \cite{kasaiNetworkVolumeAnomaly2016} & / & $651.9\pm1.3$ \\
			\midrule
			\multicolumn{3}{l}{\textbf{Proposed Methods}}\\
			\midrule
			{\tbscarlx} & 8 & $699.5\pm 1.4$ \\
			{\utbscarlx} & 8 & $706.0\pm1.8$\\
			{\autbsca} & 8  & $\mathbf{725.0\pm0.9}$\\
			{\umbscarlx} & 10 & $687.2\pm0.8$\\
			{\aumbscarlx} & 10 & $714.1\pm2.0$\\
			\midrule
			{\autbsca} (single $\tX$-Update) & 8 & $722.0\pm1.4$\\
			{\autbsca} (no augmentation) & 8 & $717.4\pm3.0$\\
			{\autbsca} (coupled weights) & 8 & $723.1\pm2.6$\\
			\midrule
		\end{tabularx}
	\end{table}
	We compare configurations of the proposed methods and reference methods in Tab.~\ref{tab:comp_synthetic}.
	{\aubsca} \cite{schynolDeepUnrollingAnomaly2023} achieves its best result for $L=7$.%
	\par
	The proposed methods outperform the reference methods in terms of \gls{auc}.
	The tensor-based architectures {\utbsca}, {\utbscarlx} and {\autbsca} leverage the low-rank structure of the data set more effectively, which is expected.
	The {\autbsca} model without leveraging the augmentation in Sec.~\ref{sec:bsca_algs}-\ref{sec:bsca_tensor_rlx} leads to a decreased validation score, which agrees with the results in Sec.~\ref{sec:simulations}-\ref{sec:sims_classical}.
	{\aumbscarlx} outperforms {\aubsca} due to its adaptive design which can take advantage of the individual statistics of flows across time and links.
	{\autbsca} achieves the highest \gls{auc} on the synthetic data $S2$ since it leverages both the low-rank \gls{cpd} model, as compared to {\aumbscarlx}, and its adaptation capability, as compared to {\utbscarlx}.
	\par
	Lastly, we consider {\autbsca} with parameter weights $\btheta\idx{\ell}$ coupled across layers.
	Specifically, we couple only the weights of the adaptive parameters, \ie, $\btheta_{\tW}\idx{\ell} = \btheta_{\tW}$ and $\btheta_{\tM}\idx{\ell} = \btheta_{\tM}$ for all $\ell = 1, \dots, L$.
	If the layer learns adaptation to the state \wrt the optimization trajectory, the gradients during training can be denoised and the data efficiency improved without significantly impacting the performance \cite{chowdhuryDeepGraphUnfolding2024}.
	Note that the non-adaptive parameters $\lambda\idx{\ell}$ are $\nu\idx{\ell}$ remain decoupled. %
	Indeed, Tab.~\ref{tab:comp_synthetic} shows that the decrease in performance is moderate for the trade-off of fewer learnable weights and significantly stabilized training.

	\subsubsection{Domain Adaptation}
	\begin{table}[t]
		\centering
		\caption{Estimated validation \gls{auc} of {\autbsca} model when trained and validated on data sets of different flow graph sizes. \label{tab:graphsize}}
		\begin{tabular}{c|c|ccc}
			\toprule
			\multicolumn{2}{c}{\multirow{2}{*}{\makecell{\\[2mm]AUC  / $10^{-3}$}}} &\multicolumn{3}{c}{\textbf{Validation}}\\
			\cmidrule{3-5}
			\multicolumn{2}{c}{} & \makecell{\\[-4mm]$N=8$\\$E=12$} & \makecell{\\[-4mm]$N=16$\\$E=32$} & \makecell{\\[-4mm]$N=32$\\$E=90$}\\
			\cmidrule{3-5}
			\multirow{3}{*}{\rotatebox[origin=c]{90}{\textbf{Training~~~~~}}} & \makecell{$N=8$\\$E=12$} & $771.8\pm5.5$ & $694.6\pm9.1$  & $617.4\pm27.2$  \\[3mm]
			& \makecell{$N=16$\\$E=32$} & $\mathbf{766.4\pm1.5}$  & $\mathbf{716.0\pm9.1}$  & $\mathbf{666.1\pm1.3}$  \\[3mm]
			& \makecell{$N=32$\\$E=90$} & $734.7\pm7.5$  & $706.4\pm5.8$  & $671.8\pm1.0$ \\
			\bottomrule
		\end{tabular}
	\end{table}
	In the following, the domain adaptation and generalization of the {\autbsca} models in case of substantial changes in flow graph size and graph connectivity is investigated.
	For this purpose, three additional synthetic data sets similar to S2 are considered, which differ in the number of nodes $N$ and directional edges $E$.
	Note that the density $\frac{E}{N(N-1)}$ of the graphs varies between the data sets as well.
	\par
	In Tab.~\ref{tab:graphsize} {\autbsca} models, that are trained based on a particular graph size and then validated on all three graph sizes, are considered.
	As expected, the best performing model for each network graph size is trained with data of the respective size.
	The generalization of the model trained on the intermediate size (highlighted) is excellent with only a decrease of approx\onedot~$0.005$ in the validation score compared to the best score.
	The model trained on the largest graph size generalizes well to the intermediate data set, however, the score reduces for smaller graphs. 
	
	\subsubsection{Computational Cost}
	\begin{table}[t]
		\centering
		\caption{Average online computation time per scenario of S2.. \label{tab:computationalcost}}
		\begin{tabularx}{\linewidth}{cXr}
			\toprule
			& \textbf{Architecture} & \textbf{Computation Cost} [\unit{\milli\second}]\\
			\midrule
			\multirow{3}{*}{\rotatebox[origin=c]{90}{\textbf{PCA}}} & {\bbcd} \cite{mardaniDynamicAnomalographyTracking2013} & $3489\pm422$\\
			& {\aubsca} \cite{schynolDeepUnrollingAnomaly2023} & $149\pm1$\\
			& {\aumbscarlx} & $125\pm1$\\
			\midrule
			\multirow{4}{*}{\rotatebox[origin=c]{90}{\textbf{CPD}}} & {\kasai} \cite{kasaiNetworkVolumeAnomaly2016} & $(114.7\pm 0.9)\cdot10^3$\\
			& {\tbscarlx} & $2544\pm341$ \\
			& {\utbscarlx} & $730\pm39$ \\
			& {\autbsca} & $773\pm33$\\
			& {\autbsca} (no augmentation) & $4150\pm93$ \\
			\midrule
		\end{tabularx}
	\end{table}
	In Tab.~\ref{tab:computationalcost}, the computational cost of the methods during validation is compared on the data set $S2$.
	The measurements are performed on a desktop with a Ryzen 2700X CPU and \qty{32}{\giga\byte} memory.
	For comparability, we choose $L=8$ layers except for {\kasai}, which uses a Hankel window size of $2$.
	The computational cost of the methods is significantly affected by the particular choice of the maximum factorization model rank $R$ or $\cprank$.
	Since this choice substantially differs for $S2$ with $R=60$ for matrix-factorization-based methods and $\cprank=300$ for tensor-based methods,  we focus on the comparison within these two groups.
	\par
	Considering the tensor-based methods, {\kasai} exhibits a substantially higher computational cost when applied along the entire sequence.
	{\autbsca} significantly reduces the computational cost compared to {\autbsca} without the augmentation approach.
	We observe a similar reduction in computational cost for the matrix-factorization based {\aumbscarlx} method compared to {\aubsca}.
	The proposed adaptive parameters of {\autbsca} incur an additional computational cost of approx\onedot $6\%$ compared to {\utbscarlx}.
	The high computational cost and large variance in runtime of {\bbcd} and {\tbscarlx} is explained by the larger number of iterations, which is significantly higher than the number of unrolled model layers, and different numbers of iterations for each data split.
	\subsubsection{Comparison on Real World Data}\label{sec:simulations_unrolled_rw}
	\begin{table}[t]
		\caption{Estimated validation \gls{auc} on the real-world data set RW.\label{tab:rw_comp}}
		\begin{tabularx}{\linewidth}{Xrr}
			\toprule
			\textbf{Architecture} & $L$ & AUC / $10^{-3}$\\
			\midrule
			{\bbcd} \cite{mardaniDynamicAnomalographyTracking2013} &  / & $778.0\pm 8.2$\\
			{\aubsca} \cite{schynolDeepUnrollingAnomaly2023} & $7$& $788.5\pm 4.8$\\
			{\kasai} \cite{kasaiNetworkVolumeAnomaly2016} & / & $712.4\pm 18.8$\\
			{\aumbscarlx} (prop.) & $7$ & $797.4\pm12.7$\\
			{\autbsca} (prop.) & $9$ & $\mathbf{803.9\pm12.0}$\\
			\midrule
		\end{tabularx}
	\end{table}
	\begin{figure}
		\centering
\tikzsetnextfilename{main_rw_roc}
\begin{tikzpicture}
\pgfplotsset{
	layers/my layer set/.define layer set={
		background,
		main,
		foreground,
	}{
	},
	set layers=my layer set,
}
\begin{axis}[
	xmin=0, xmax=1,
	ymin = 0, ymax = 1,
	width = 6.5cm,
	height = 6.5cm,
	minor tick num = 0,
	xlabel near ticks,
	ylabel near ticks,
	xlabel = {$P_\upm{False Alarm}$},
	ylabel = {$P_\upm{Detection}$},
	mark repeat  = 10,
	mark phase = 5,
	legend style={legend columns=3, at={(1.2,1.03)}},
	]
	\addplot+[red, solid, mark phase = 2] table[col sep = tab, x index = 0, y index = 1]{data/rw/BSCAUnrolled_abilene15_cb2_raw_mxmul0.5_ly7_r30_paramnw_aucfs_roc_combined.txt}; %
	\addplot+[orange, solid, mark phase = 6] table[col sep = tab, x index = 0, y index = 1]{data/rw/BSCATensorUnrolled_abilene15_cb2_raw_mxmul0.5_ly7_r30_notens_mw_dff1lymuf1ly_rlx2x+nnf_it1nrlx_aucfs_roc_combined.txt}; %
	\addplot+[purple, solid, mark phase = 10] table[col sep = tab, x index = 0, y index = 1]{data/rw/refalg_kasai_r420_wx_ON_abilene15_cb2_raw_mxmul0.5_roc_combined.txt}; %
	\addplot+[blue, solid, mark phase = 4] table[col sep = tab, x index = 0, y index = 1]{data/rw/bayopt_bbcd_r30_on_abilene15_cb2_raw_mxmul0.5_roc_combined.txt}; %
	\addplot+[green, solid, mark phase = 8] table[col sep = tab, x index = 0, y index = 1]{data/rw/BSCATensorUnrolled_abilene15_cb2_raw_mxmul0.5_ly9_r420_mw_dff1lymuf1ly_rlx2x+nnf_it1nrlx_aucfs_roc_combined.txt}; %
	\pgfplotsforeachungrouped \i/\j in {5/6,7/8,9/10,11/12,13/14,15/16,17/18,19/20,21/22,23/24,25/26,27/28}
	{
		\addplot+[red, solid, thick, mark=none, opacity=0.1, on layer=background] table[col sep = tab, x index =\i, y index = \j] {data/rw/BSCAUnrolled_abilene15_cb2_raw_mxmul0.5_ly7_r30_paramnw_aucfs_roc_combined.txt}; %
	}
	\pgfplotsforeachungrouped \i/\j in {5/6,7/8,9/10,11/12,13/14,15/16,17/18,19/20,21/22,23/24,25/26,27/28}
	{
		\addplot+[blue, solid, thick, mark=none, opacity=0.1, on layer=background] table[col sep = tab, x index =\i, y index = \j] {data/rw/bayopt_bbcd_r30_on_abilene15_cb2_raw_mxmul0.5_roc_combined.txt}; %
	}
	\pgfplotsforeachungrouped \i/\j in {5/6,7/8,9/10,11/12,13/14,15/16,17/18,19/20,21/22,23/24,25/26,27/28}
	{
		\addplot+[orange, solid, thick, mark=none, opacity=0.1, on layer=background] table[col sep = tab, x index =\i, y index = \j] {data/rw/BSCATensorUnrolled_abilene15_cb2_raw_mxmul0.5_ly7_r30_notens_mw_dff1lymuf1ly_rlx2x+nnf_it1nrlx_aucfs_roc_combined.txt}; %
	}
	\pgfplotsforeachungrouped \i/\j in {5/6,7/8,9/10,11/12,13/14,15/16,17/18,19/20,21/22,23/24,25/26,27/28}
	{
		\addplot+[green, solid, thick, mark=none, opacity=0.1, on layer=background] table[col sep = tab, x index =\i, y index = \j] {data/rw/BSCATensorUnrolled_abilene15_cb2_raw_mxmul0.5_ly9_r420_mw_dff1lymuf1ly_rlx2x+nnf_it1nrlx_aucfs_roc_combined.txt}; %
	}
	\pgfplotsforeachungrouped \i/\j in {5/6,7/8,9/10,11/12,13/14,15/16,17/18,19/20,21/22,23/24,25/26,27/28}
	{
	\addplot+[purple, solid, thick, mark=none, opacity=0.1, on layer=background] table[col sep = tab, x index =\i, y index = \j] {data/rw/refalg_kasai_r420_wx_ON_abilene15_cb2_raw_mxmul0.5_roc_combined.txt}; %
	}
	\coordinate (inset) at (axis cs:0.94,0.1);
	\coordinate (zoom00) at (axis cs:0.0,0.0);
	\coordinate (zoom10) at (axis cs:0.09,0.0);
	\coordinate (zoom01) at (axis cs:0.0,0.4);
	\coordinate (zoom11) at (axis cs:0.09,0.4);
	\node[rectangle, darkGray, inner sep=0mm, draw, dashed, line width=0.25mm, fit=(zoom00)(zoom01)(zoom11)(zoom10)] {};
	
	\legend{{\aubsca}, {\aumbscarlx} (prop.), {\bbcd}, {\autbsca} (prop.), {\kasai}}
	\legend{{\aubsca}, {\aumbscarlx} (prop.), {\kasai}, {\bbcd}, {\autbsca} (prop.)}
\end{axis}
\begin{axis}[
	at={(inset)},
	anchor=south east, 
	axis background/.style={fill=gray!10},
	xmin=0, xmax=0.09,
	ymin = 0, ymax = 0.4,
	width = 4cm,
	height = 4cm,
	minor tick num = 0,
	xtick={0, 0.03, 0.06, 0.09},
	xticklabels = {0, 0.03, 0.06, 0.09},
	xticklabel style={scaled x ticks=false},
	xlabel near ticks,
	ylabel near ticks,
	mark repeat  = 10,
	mark phase = 5,
	]
	\pgfplotsforeachungrouped \i/\j in {5/6,7/8,9/10,11/12,13/14,15/16,17/18,19/20,21/22,23/24,25/26,27/28}
	{
		\addplot+[red, solid, thick, mark=none, opacity=0.1, on layer=main] table[col sep = tab, x index =\i, y index = \j] {data/rw/BSCAUnrolled_abilene15_cb2_raw_mxmul0.5_ly7_r30_paramnw_aucfs_roc_combined.txt}; %
	}
	\pgfplotsforeachungrouped \i/\j in {5/6,7/8,9/10,11/12,13/14,15/16,17/18,19/20,21/22,23/24,25/26,27/28}
	{
		\addplot+[blue, solid, thick, mark=none, opacity=0.1, on layer=main] table[col sep = tab, x index =\i, y index = \j] {data/rw/bayopt_bbcd_r30_on_abilene15_cb2_raw_mxmul0.5_roc_combined.txt}; %
	}
	\pgfplotsforeachungrouped \i/\j in {5/6,7/8,9/10,11/12,13/14,15/16,17/18,19/20,21/22,23/24,25/26,27/28}
	{
		\addplot+[orange, solid, thick, mark=none, opacity=0.1, on layer=main] table[col sep = tab, x index =\i, y index = \j] {data/rw/BSCATensorUnrolled_abilene15_cb2_raw_mxmul0.5_ly7_r30_notens_mw_dff1lymuf1ly_rlx2x+nnf_it1nrlx_aucfs_roc_combined.txt}; %
	}
	\pgfplotsforeachungrouped \i/\j in {5/6,7/8,9/10,11/12,13/14,15/16,17/18,19/20,21/22,23/24,25/26,27/28}
	{
		\addplot+[green, solid, thick, mark=none, opacity=0.1, on layer=main] table[col sep = tab, x index =\i, y index = \j] {data/rw/BSCATensorUnrolled_abilene15_cb2_raw_mxmul0.5_ly9_r420_mw_dff1lymuf1ly_rlx2x+nnf_it1nrlx_aucfs_roc_combined.txt}; %
	}
	\pgfplotsforeachungrouped \i/\j in {5/6,7/8,9/10,11/12,13/14,15/16,17/18,19/20,21/22,23/24,25/26,27/28}
	{
		\addplot+[purple, solid, thick, mark=none, opacity=0.1, on layer=main] table[col sep = tab, x index =\i, y index = \j] {data/rw/refalg_kasai_r420_wx_ON_abilene15_cb2_raw_mxmul0.5_roc_combined.txt}; %
	}
	\addplot+[red, solid, mark phase = 2, on layer=main] table[col sep = tab, x index = 0, y index = 1]{data/rw/BSCAUnrolled_abilene15_cb2_raw_mxmul0.5_ly7_r30_paramnw_aucfs_roc_combined.txt}; %
	\addplot+[blue, solid, mark phase = 4, on layer=main] table[col sep = tab, x index = 0, y index = 1]{data/rw/bayopt_bbcd_r30_on_abilene15_cb2_raw_mxmul0.5_roc_combined.txt}; %
	\addplot+[orange, solid, mark phase = 6, on layer=main] table[col sep = tab, x index = 0, y index = 1]{data/rw/BSCATensorUnrolled_abilene15_cb2_raw_mxmul0.5_ly7_r30_notens_mw_dff1lymuf1ly_rlx2x+nnf_it1nrlx_aucfs_roc_combined.txt}; %
	\addplot+[green, solid, mark phase = 8, on layer=main] table[col sep = tab, x index = 0, y index = 1]{data/rw/BSCATensorUnrolled_abilene15_cb2_raw_mxmul0.5_ly9_r420_mw_dff1lymuf1ly_rlx2x+nnf_it1nrlx_aucfs_roc_combined.txt}; %
	\addplot+[purple, solid, mark phase = 10, on layer=main] table[col sep = tab, x index = 0, y index = 1]{data/rw/refalg_kasai_r420_wx_ON_abilene15_cb2_raw_mxmul0.5_roc_combined.txt}; %
	\coordinate (inset00) at (axis cs:0.0,0.0);
	\coordinate (inset10) at (axis cs:0.09,0.0);
	\coordinate (inset01) at (axis cs:0.0,0.4);
	\coordinate (inset11) at (axis cs:0.09,0.4);
\end{axis}
\draw[dashed, darkGray, line width=0.1mm] (zoom00) to (inset00);
\draw[dashed, darkGray, line width=0.1mm] (zoom01) to (inset01);
\draw[dashed, darkGray, line width=0.1mm] (zoom10) to (inset10);
\draw[dashed, darkGray, line width=0.1mm] (zoom11) to (inset11);
\end{tikzpicture}
		\caption{Averaged \gls{roc} curves (solid) and per-scenario \gls{roc} curves (semi-transparent) achieved by several methods on real-world data. \label{fig:rw_roc}}
	\end{figure}
	In Tab.~\ref{tab:rw_comp}, {\utbscarlx}, {\autbsca} and {\aumbscarlx} are compared against {\kasai} and {\aubsca} on the real-world data set RW.
	Where applicable, the number of layers achieving the best validation score is chosen.
	This data set poses a particular challenge to learning-based methods since the number of available scenarios is very low. %
	In addition, the link load amplitudes vary by orders of magnitude from link to link.
	\par
	{\aumbscarlx} and {\autbsca} outperform {\aubsca} and {\bbcd}, whereas the score achieved by {\kasai} is substantially lower.
	The per-slice features and adaptivity of {\aumbscarlx} and {\autbsca} lead to an improvement in performance due to the different magnitudes of flows across links and time.
	However, {\aumbscarlx} and {\autbsca} achieve a similar estimated \gls{auc}. %
	\par
	The \gls{roc} curve of the same experiments is shown in Fig.~\ref{fig:rw_roc}.
	{\autbsca} exhibits a significantly steeper slope at the origin on average, \ie, the detection performance is superior if the probability of false alarms is low, which is typically the useful region of operation.
	This indicates that {\autbsca} indeed leverages periodicity in the real-world data sequence at the cost of computational complexity
	
	\section{Conclusion and Outlook}\label{sec:conclusion}
	\Gls{ad} in network flows is tackled using a low-rank tensor \gls{cpd} model, sparsity and algorithm unrolling. 
	Based on block-successive convex approximations, convergent algorithms are proposed to recover anomalies from a batch of possibly incomplete observations.
	We introduce an augmentation to substantially reduce the computational complexity of the algorithm.
	The proposed algorithms are extended into lightly parametrized \gls{dn} architectures, which adapt to flow statistics across origin-destination pairs and points in time using permutation equivariant feature embeddings and learnable functions of statistical parameters.
	The permutation equivariance of the methods allows for domain adaptation to arbitrary network graphs, routing and sequence lengths.
	We further propose an approximate and soft \gls{auc} estimation loss to accelerate the training and finetune the model weights to the \gls{ad} task.
	Using extensive experiments on synthetic data, the data efficiency of the proposed architecture, adaptation capability and effectiveness on periodic flows with varying statistics is verified.
	We further demonstrate superior detection performance compared to reference algorithms in a real-world data set with a small number of training examples.
	\par
	Future work may consider improving feature embeddings by graph-based embeddings across the flow dimension, shift-invariant convolutional embeddings across the time, the incorporation of additional structural information of anomalies similar to \cite{yeAnomalyTolerantTrafficMatrix2017}, and investigations into efficient semi-supervised or unsupervised loss functions.

	\appendices

	\section{} \label{sec:apx_auc_label_err}
	Define $\auc(\bA; \setestAgt_1, \setestAgt_0)$ as the computable \gls{auc} estimate based on the available labels and $\auc(\bA; \setestAgt_1, \setAgt_0)$ as the AUC estimate based on the labeled anomalies and the true non-anomalies.
	We can write $\auc(\bA; \setestAgt_1, \setestAgt_0)$ as
	\begin{align}
		&\auc(\bA; \setestAgt_1, \setestAgt_0) \nonumber\\
		&\quad=\frac{1}{\abs{\setestAgt_1}\abs{\setestAgt_0}}
		\Bigg( 
		\underbrace{\sum_{\substack{\forall(i_1, t_1)\\\in \setestAgt_1(\ssample)}} \sum_{\substack{\forall(i_0, t_0)\\\in \setAgt_0(\ssample)}} \stepfun\left(\matel{\hbA}{i_1,t_1} \!\!\!\!\!\! - \matel{\hbA}{i_0,t_0}\right)}_{=\abs{\setestAgt_1}\abs{\setAgt_0}\auc(\bA; \setestAgt_1, \setAgt_0)} \nonumber\displaybreak[1]\\
		&\qquad+ \sum_{\substack{\forall(i_1, t_1)\\\in \setAgt_1(\ssample)}} \sum_{\substack{\forall(i_0, t_0)\\\in \setAul_1(\ssample)}} \stepfun\left(\matel{\hbA}{i_1,t_1} \!\!\!\!\!\!- \matel{\hbA}{i_0,t_0}\right)\Bigg)
	\end{align}
	where $\setestAgt_0 = \setAgt_0 \cup \setAul_1$.
	Since $\abs{\setAul_1} = \abs{\setestAgt_0} - \abs{\setAgt_0}$, we have
	\begin{align}
		&\abs{\auc(\bA; \setestAgt_1, \setestAgt_0) - \auc(\bA; \setestAgt_1, \setAgt_0)} \nonumber\\
		&\quad=\Bigg\vert\frac{1}{\abs{\setestAgt_1}\abs{\setestAgt_0}}\sum_{\substack{\forall(i_1, t_1)\\\in \setAgt_1(\ssample)}} \sum_{\substack{\forall(i_0, t_0)\\\in \setAul_1(\ssample)}} \stepfun\left(\matel{\hbA}{i_1,t_1} \!\!\!\!\!\!- \matel{\hbA}{i_0,t_0}\right) \nonumber\displaybreak[1]\\
		&\qquad- \frac{\abs{\setAul_1}}{\abs{\setestAgt_0}}\auc(\bA; \setestAgt_1, \setAgt_0)\Bigg\vert\displaybreak[1]\\
		&\quad= \frac{\abs{\setAul_1}}{\abs{\setestAgt_0}} \Bigg\vert \!\underbrace{\frac{1}{\abs{\setestAgt_1}\abs{\setAul_1}}\! \sum_{\substack{\forall(i_1, t_1)\\\in \setAgt_1(\ssample)}} \sum_{\substack{\forall(i_0, t_0)\\\in \setAul_1(\ssample)}}\!\!\!\! \stepfun\left(\matel{\hbA}{i_1,t_1} \!\!\!\!\!\!- \matel{\hbA}{i_0,t_0}\right)}_{=\auc(\bA; \setestAgt_1, \setAul_1) \in \bmat{0, 1}} \nonumber\\
		&\qquad- \underbrace{\auc(\bA; \setestAgt_1, \setAgt_0)}_{\in\bmat{0,1}} \Bigg\vert \leq \frac{\abs{\setAul_1}}{\abs{\setestAgt_0}}.
	\end{align}

	\bibliographystyle{IEEEtran}
	\bibliography{unrolled_nw_anomaly}
	\begin{IEEEbiography}[{\includegraphics[width=1in,height=1.25in,clip,keepaspectratio]{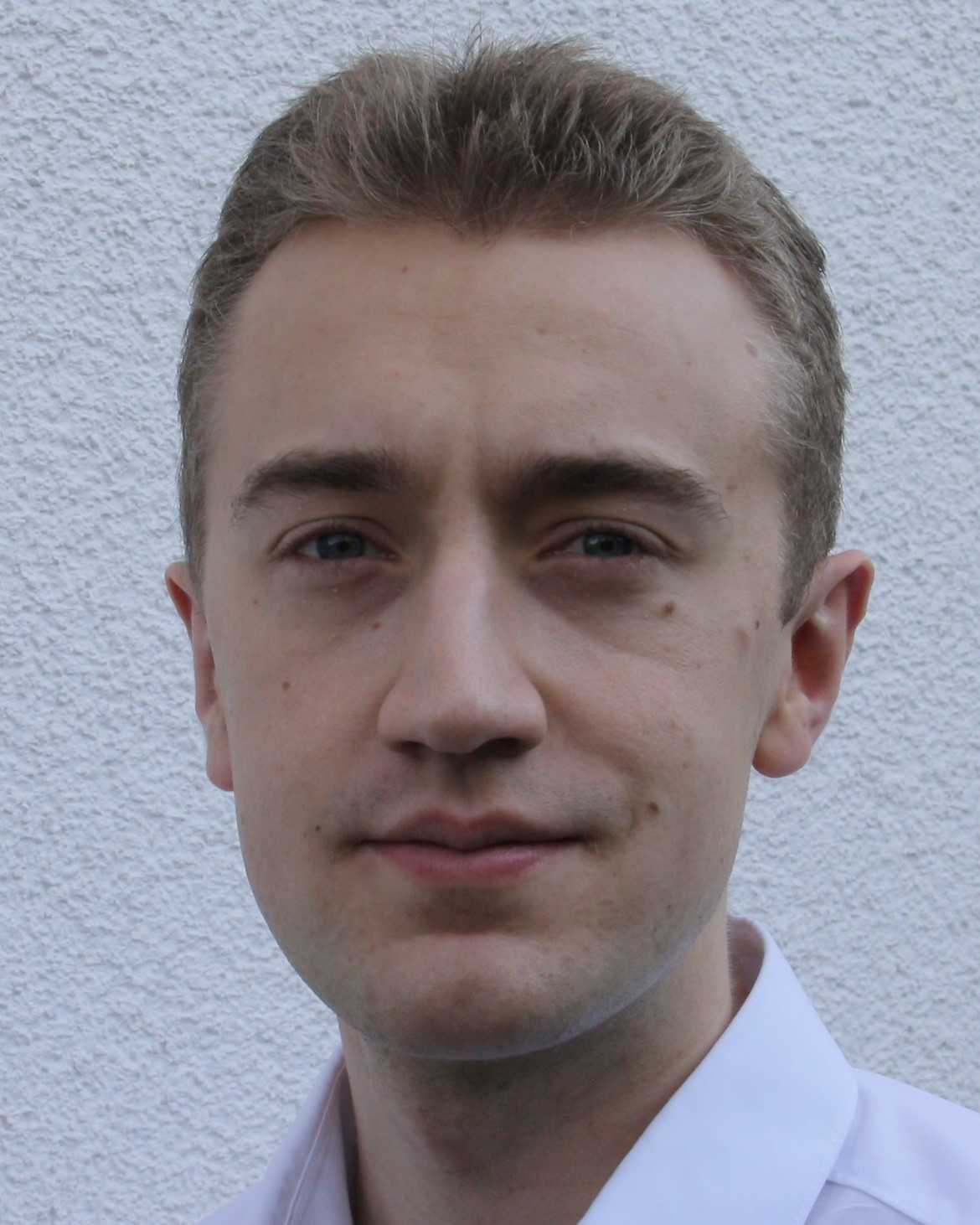}}]
		{Lukas Schynol}, received the M.Sc. degree in 2021 in electrical engineering from Technische Universität Darmstadt, Darmstadt, Germany, where he is currently working towards the Ph.D. degree. His research interests include resource allocation and anomaly detection in wireless networks using model-assisted deep learning. He received the best student paper award (2nd place) at the IEEE International Workshop on Computational Advances in Multi-Sensor Adaptive Processing (CAMSAP) in 2023.
	\end{IEEEbiography}
	\begin{IEEEbiography}[{\includegraphics[width=1in,height=1.25in,clip,keepaspectratio]{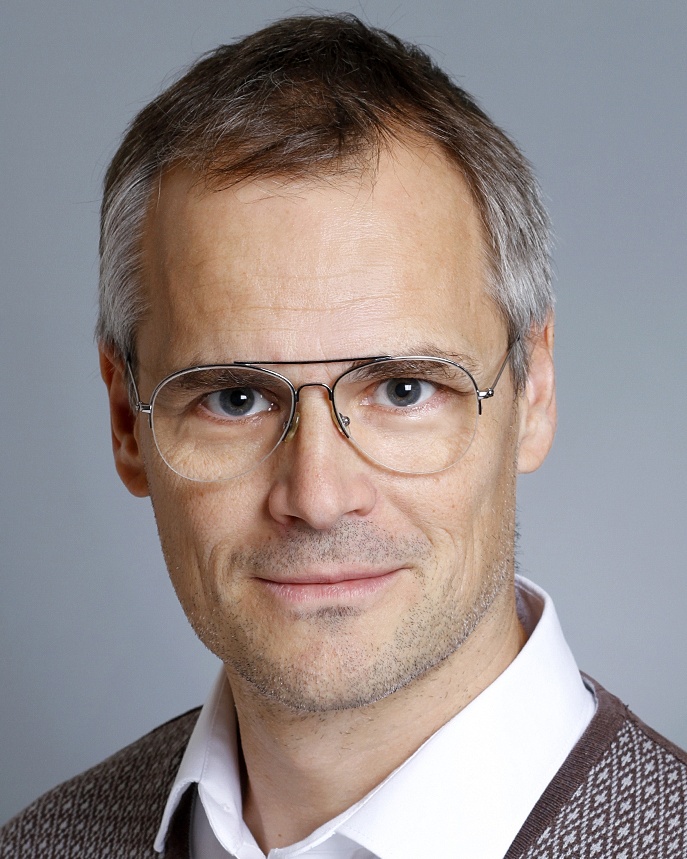}}]
		{Marius Pesavento}, received the Dipl.-Ing. and the PhD degree from Ruhr-Universität Bochum, Germany, in 1999 and 2005, respectively, and the M.Eng. degree from McMaster University, Hamilton, ON, Canada, in 2000. From 2005 to 2008, he was a Research Engineer with two startup companies. He became an Assistant Professor in 2010 and a Full Professor of Communication Systems in 2013 with the Department of Electrical Engineering and Information Technology, Technische Universität Darmstadt, Germany. His research interests include statistical and sensor array signal processing, MIMO communications, optimization and model assisted deep learning. He is Vice-Chair of the IEEE SPS “Sensor Array and Multichannel Signal Processing” Technical Committee (member 2012–2017), Member of the “Signal Processing Theory and Methods” Technical Committee (since 2021) and Past Chair of the EURASIP “Signal Processing for Multisensor Systems” Technical Area Committee of the European Association for Signal Processing (EURASIP). He is a Deputy Editor-in-Chief of the IEEE Open Journal of Signal Processing (SAE 2019–2024) and was Associate Editor of the IEEE Transactions on Signal Processing (2012–2016), and Subject Editor of the EURASIP journal Signal Processing (since 2024, Handling Editor 2011–2023). He is the Regional Director-at-Large for Region 8 (2025–2027).
	\end{IEEEbiography}
	\vfill\pagebreak
\end{document}